\pdfoutput=1

\documentclass[twocolumn,11pt]{article}
\usepackage{tikz}
\usetikzlibrary{backgrounds}

\usepackage[final]{acl}
\newcounter{prompt}
\usepackage{times}
\usepackage{makecell}
\usepackage{latexsym}
\usepackage{enumitem}
\usepackage{array}
\usepackage{algorithm}
\usepackage{algpseudocode}
\usepackage{amsmath}
\usepackage{booktabs}
\usepackage{cuted}
\usepackage{pifont}      
\usepackage{xcolor}      
\usepackage{colortbl}    
\usepackage{hyperref}    
\definecolor{LightGray}{gray}{0.93}
\definecolor{BluePrompt}{RGB}{111, 168, 220}
\definecolor{GreenPrompt}{RGB}{0, 150, 0}
\definecolor{PurplePrompt}{RGB}{142, 124, 195}
\definecolor{OrangePrompt}{RGB}{234, 122, 6}
\definecolor{PinkPrompt}{RGB}{234, 130, 184}
\definecolor{YellowPrompt}{RGB}{239, 183, 12}
\definecolor{TealPrompt}{RGB}{118, 165, 175}
\definecolor{RedPrompt}{RGB}{224, 102, 102}
\usepackage{listings}  
\newcommand{\cmark}{\textcolor{green!60!black}{\ding{51}}}
\newcommand{\xmark}{\textcolor{red!70!black}{\ding{55}}}

\lstdefinelanguage{json}{
    basicstyle=\ttfamily\small,
    showstringspaces=false,
    breaklines=true,
    morestring=[b]",
    morecomment=[l]{//},
    morecomment=[s]{/*}{*/},
    stringstyle=\ttfamily,
}

\usepackage[T1]{fontenc}

\usepackage[utf8]{inputenc}

\usepackage{microtype}

\usepackage{inconsolata}

\usepackage{graphicx}
\usepackage{lipsum}

\usepackage[most]{tcolorbox}
\tcbset{
  promptbox/.style={
    colback=gray!10,
    colframe=gray,
    coltitle=white,
    fonttitle=\bfseries,
    boxrule=0.5pt,
    arc=2mm,
    boxsep=5pt,
    left=4pt, right=4pt, top=4pt, bottom=4pt,
    enhanced
  }
}

\usepackage{cuted} 

\tcbset{
  algobox/.style={
    colback=white,
    colframe=gray,
    fonttitle=\bfseries,
    title=Algorithm 1: Agentic Neural Network with Dynamic Routing and Adaptive Optimization,
    boxrule=0.4pt,
    arc=1mm,
    boxsep=4pt,
    left=4pt, right=4pt, top=4pt, bottom=4pt,
    enhanced
  }
}

\title{Self-Evolving Multi-Agent Systems via Textual Backpropagation}

\author{
Xiaowen Ma\textsuperscript{1}\thanks{Equal contribution.}\thanks{Corresponding author: \textit{maxiaowen0929@gmail.com, cognitive.yunpu@gmail.com}}  
Yunpu Ma\textsuperscript{1,3}\footnotemark[1]\footnotemark[2]
 \textbf{Chenyang Lin\textsuperscript{2}} \quad
 \textbf{Sikuan Yan\textsuperscript{1,3}} \quad
 \textbf{Jinhe Bi\textsuperscript{1,3}} \\
 \textbf{Zixuan Cao\textsuperscript{1}} \quad
 \textbf{Yijun Tian\textsuperscript{4}} \quad
 \textbf{Volker Tresp\textsuperscript{1,3}}  \quad
 \textbf{Hinrich Schuetze\textsuperscript{1,3}}
\\
\\
 \textsuperscript{1}Ludwig Maximilian University of Munich \quad
\textsuperscript{2}Technical University of Munich \\
 \textsuperscript{3}Munich Center for Machine Learning \quad
 \textsuperscript{4}University of Notre Dame
}

\begin{document}
\maketitle
\begin{abstract}

Leveraging multiple Large Language Models (LLMs) has proven effective for addressing complex, high-dimensional tasks, but current approaches often rely on static, manually engineered multi-agent configurations. To overcome these constraints, we present the Agentic Neural Network ($\mathcal{ANN}$), a framework that conceptualizes multi-agent collaboration as a layered neural network architecture. In this design, each agent operates as a node, and each layer forms a cooperative team focused on a specific subtask. Agentic Neural Network follows a two-phase optimization strategy: (1) Forward Phase - Drawing inspiration from neural network forward passes, tasks are dynamically decomposed into subtasks, and cooperative agent teams with suitable aggregation methods are constructed layer by layer. (2) Backward Phase - Mirroring backpropagation, we refine both global and local collaboration through iterative feedback, allowing agents to self-evolve their roles, prompts, and coordination. This neuro-symbolic approach enables $\mathcal{ANN}$ to create new or specialized agent teams post-training, delivering notable gains in accuracy and adaptability. Across seven benchmark datasets, $\mathcal{ANN}$ surpasses leading multi-agent baselines under the same configurations, showing consistent performance improvements.  

\end{abstract}

\section{Introduction}

\begin{figure*}[t]
  \includegraphics[width=0.48\linewidth]{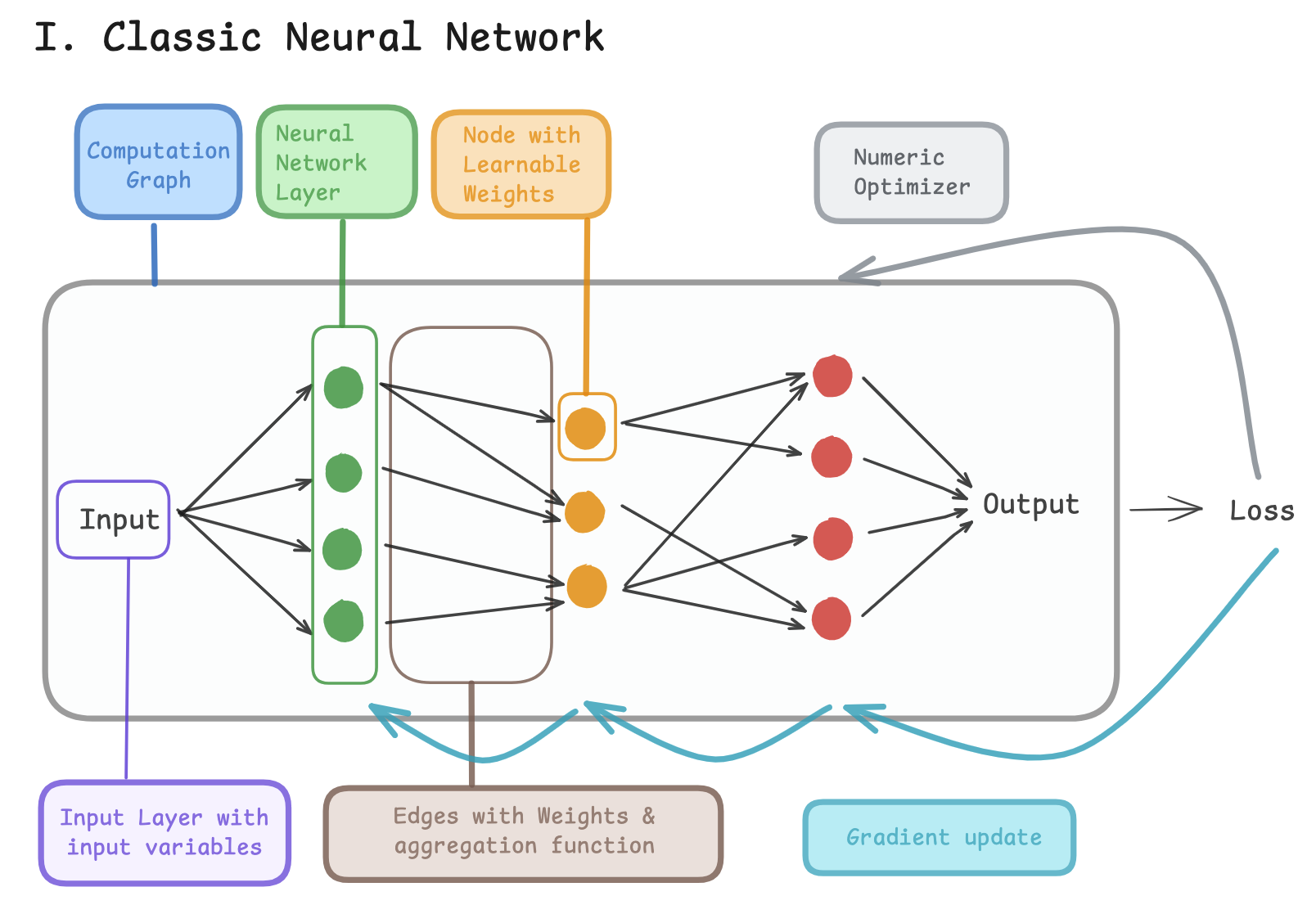}  \hfill
  \includegraphics[width=0.48\linewidth]{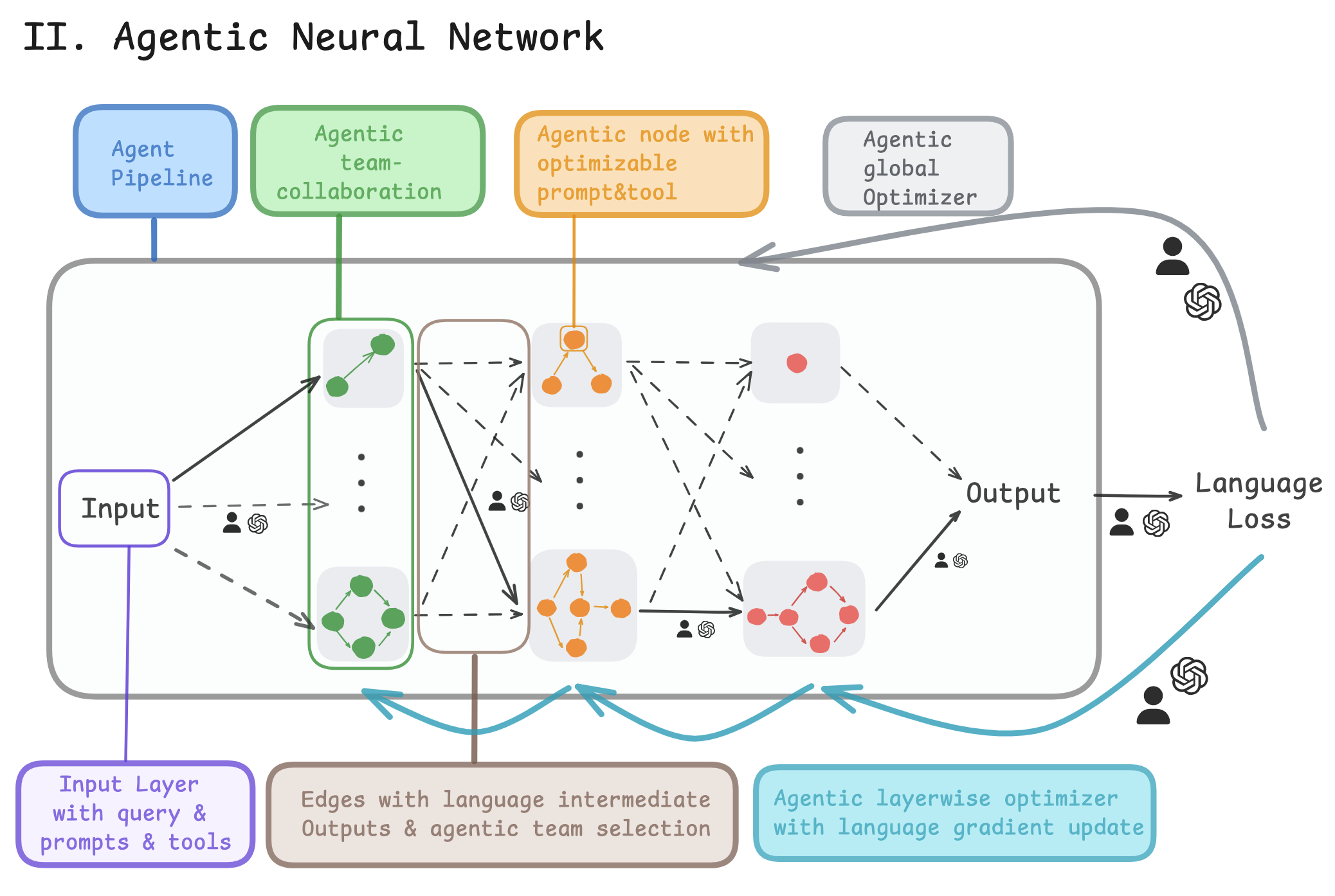}
    \caption{
    A conceptual comparison between classic neural networks (left) and our $\mathcal{ANN}$ (right). In the right-hand agentic diagram, the brown module labeled \emph{“Edges with language intermediate Outputs \& agentic team selection”} represents the choice among multiple candidate collaboration strategies between agent teams. \textbf{Solid lines} indicate selected collaboration modes that form the pipeline connection between layers, while \textbf{dashed lines} represent alternative strategies that were not selected at that step.}
  \label{fig:classic-agentic}
\end{figure*}

Large Language Models (LLMs) have ushered in a new era of artificial intelligence, exhibiting strong capabilities in reasoning, content generation, and multi-step problem-solving \cite{kojima2022,Ouyang2022TrainingLM}. By grouping these models into \emph{multi-agent systems} (MAS), researchers have addressed an array of complex tasks, ranging from code generation and debugging \cite{jimenez2023} to retrieval-augmented generation \cite{Khattab2023DSPyCD,Lewis2020RetrievalAugmentedGF,Gao2023RetrievalAugmentedGF} and data analysis \cite{Hong2024DataIA,Hu2024InfiAgentDABenchEA}. Often, MAS outperform their single-agent equivalents by bringing together diverse agent roles and expertise, including verifier agents \cite{Shinn2023ReflexionLA} or debating agents \cite{Qian2024ScalingLM,Zhuge2024LanguageAA}, thus creating more adaptable and robust solutions. However, designing and deploying effective MAS remains demanding. Developers frequently invest substantial effort into prompt engineering, role assignment, and topology definition by trial and error \cite{chen2024agentverse,Hong2023MetaGPTMP}, especially for dynamic, high-dimensional tasks.

Recent advances in automating aspects of MAS design aim to relieve these challenges. For instance, \citet{Khattab2024DSPyCD} introduced systematic methods for generating in-context exemplars; \citet{hu2025automated} presented a meta-agent capable of creating new topologies in code; and \citet{zhang2024b} employed Monte Carlo Tree Search to find improved workflow configurations;   \citet{ke2025maszerodesigningmultiagentsystems} proposed an automatic MAS optimization architecture under zero supervision and demonstrated significant gains. These innovations mirror earlier developments in MAS design research, where layer-wise optimization gave way to holistic, end-to-end backpropagation \cite{jacobs1991adaptive,hinton2006fast}. Similarly, \emph{symbolic} or \emph{agent-level} frameworks that model entire multi-agent pipelines as computational graphs have emerged \cite{Khattab2023DSPyCD,zhugegptswarm,zhou2024symbolic}.

Building on these insights, we introduce the \emph{Agentic Neural Network ($\mathcal{ANN}$)}, a framework that adapts principles from classic neural networks to orchestrate multiple LLM agents. As shown in Figure~\ref{fig:classic-agentic}, conventional neural networks rely on learnable weights and numeric optimizers for end-to-end training via gradient-based updates, whereas $\mathcal{ANN}$ considers each layer as a team of language agents, jointly optimizing roles, prompts, and tools through textual gradients \cite{yuksekgonul2024textgrad}. While Mixture of Experts (MoE) and Mixture of Agents (MoA) architectures aim to scale model capacity through gated expert selection within a monolithic model \cite{shazeer2017outrageouslylargeneuralnetworks,wang2024mixtureofagentsenhanceslargelanguage}, $\mathcal{ANN}$ organizes layerwise teams of language agents that collaborate through multi-step reasoning and are refined via textual gradients \cite{yuksekgonul2024textgrad}. This design enables $\mathcal{ANN}$ to support flexible, role-based agent coordination beyond the scope of numeric expert gating.

Instead of a purely engineering-driven approach, $\mathcal{ANN}$ divides a complex task into smaller sub-problems, assigning each to a layer of specialized agents, and iteratively refines both local design (i.e., agent prompts and configurations) and global coordination (i.e., inter-layer flows and topologies). Our approach proceeds in two stages. First, during the forward agent team generation phase, the main task is decomposed into subtasks, with specialized agent teams dynamically assigned layer by layer, ensuring each layer is responsible for a distinct subtask. Then, during training, if performance is suboptimal, the backward agent team optimization phase backpropagates textual feedback to isolate errors and propose targeted adjustments. These textual critiques act like gradient signals, guiding prompt updates and connection refinements \cite{yao2023react,verma2024advances,Khattab2023DSPyCD}.

To illustrate the capabilities of our framework, we evaluate $\mathcal{ANN}$ on five challenging datasets: MATH \cite{hendrycks2021measuringmathematicalproblemsolving} for mathematical reasoning, DABench \cite{Hu2024InfiAgentDABenchEA} for data analysis, Creative Writing \cite{zhou2024symbolic} for open-ended writing tasks, HumanEval \cite{chen2021evaluatinglargelanguagemodels} for code generation, and MMLU \cite{hendrycks2021measuringmassivemultitasklanguage} for multiple-choice question answering.  Our experiments show that our framework not only simplifies MAS design by automating prompt tuning, role assignment, and agent collaboration but also outperforms existing baselines in accuracy. Through this process, the framework acquires self-evolving capabilities, dynamically reconfiguring its agent teams and coordination strategies to meet the demands of novel tasks.

\section{Related Works}
In this section, we review the evolution of AI agents into LLM-based systems, discuss the emerging concept of agentic workflows, survey automated methods for optimizing agent configurations, and outline the remaining challenges in multi-agent settings.
\paragraph{Evolution of AI Agents} 
Early AI agents were highly specialized and depended chiefly on symbolic reasoning, as seen in board-game-playing systems like Chess and Go. Subsequent innovations introduced reactive and reinforcement learning agents with greater adaptability. More recently, LLM-based agents have appeared, incorporating large-scale language models \cite{Radford2018ImprovingLU,Radford2019LanguageMA,Ouyang2022TrainingLM} at their foundation. By processing natural language inputs and outputs, these agents enable more flexible, human-like interactions and reasoning.
\paragraph{LLM-Based Agentic Workflows}
Modern workflows often rely on multiple LLM invocations to address complex, multi-step tasks \cite{Wei2022ChainOT,Madaan2023SelfRefineIR,Gao2022PALPL}. In these agentic workflows, each stage or node corresponds to specific subtasks like prompt creation, tool utilization, or domain-specific strategies \cite{Hong2023MetaGPTMP,Yang2023LargeLM,Cai2023LargeLM}. Through specialized roles—including data analyzers, verifiers, or debaters—LLM-based agents can collaborate efficiently on a range of domain challenges, from code generation \cite{Hong2024DataIA,lee2023wrote} to advanced data analysis \cite{li2024tapilot}.

\paragraph{Automated Optimization Approaches}
As task workflows grow more involved, automated methods aim to minimize manual engineering. \emph{Prompt optimization} tailors textual inputs to steer LLM outputs \cite{Khattab2023DSPyCD,Zhuge2024LanguageAA}. \emph{Hyperparameter tuning} fine-tunes model parameters or scheduling \cite{saad2024archon}, and \emph{workflow optimization} revises entire computational graphs or code structures \cite{hu2025automated,zhang2024b,zhugegptswarm}. Symbolic learning frameworks \cite{Hong2024DataIA,Zhuge2024LanguageAA,zhou2024symbolic} optimize prompts, tools, and node configurations collectively, mitigating local optima that can emerge from optimizing each component independently. Furthermore, \citet{lee2025compoundaisystemsoptimization} propose a systematic taxonomy for AI systems optimization, enabling benchmarking of MAS designs and evaluation of collaborative frameworks.

\paragraph{MAS Integration and Key Challenges}
In multi-agent systems, LLMs facilitate inter-agent communication, strategic planning, and iterative task decomposition \cite{yao2023react,wang2024astuterag}. However, scaling these agents prompts concerns about computational overhead, privacy, and the opaque “black box” nature of large models \cite{liu2024,verma2024advances}. These considerations highlight the need for robust design, continuous oversight, and data-centric strategies that balance performance and interpretability.

Overall, the field has moved from manually designed agent architectures to more data-driven, automated approaches that harness LLMs’ language capabilities. Despite noteworthy gains in prompt tuning, structural optimization, and integrated workflows, a gap remains for frameworks that unify these methods into efficient, adaptable, and end-to-end automated systems suited for large-scale real-world deployments.

\begin{figure*}[t] 
\centering
  \includegraphics[width=1\linewidth]{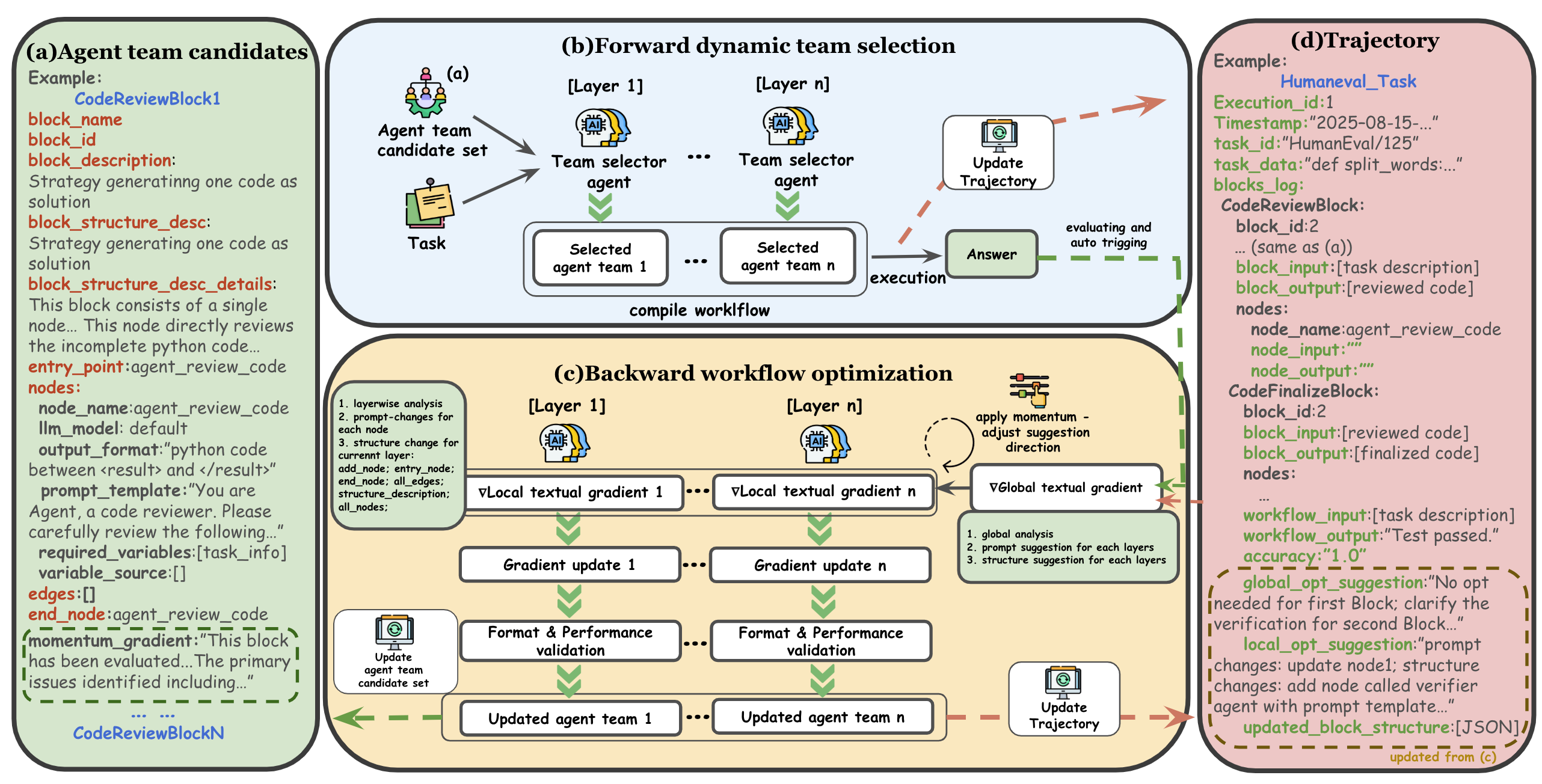} 

 

        \caption{Overview of our $\mathcal{ANN}$ framework with textual backpropagation, which dynamically selects and refines agent teams layer by layer.
(a) Agent team candidates: layer-wise candidate sets $F_\ell$ are represented as structured blocks composed of nodes, edges, and prompts.
(b) Forward dynamic team selection: a team-selector agent selects agent team for each layer, compiles the workflow, executes the task, and records the trajectory.
(c) Backward workflow optimization: task evaluation and the recorded trajectory generate a textual gradient $\nabla_{\text{text}}$, which decomposes into a global gradient for workflow structure and local gradients for updating prompts, nodes, and edges at each layer. Momentum-based updates and format/performance validation ensure stable refinement before integrating changes into the candidate pools.
(d) Trajectory: a structured record capturing the inputs and outputs of each node, layer, and workflow, along with global and local suggestions that not only guide agent team updates but are also stored as momentum gradients within the corresponding blocks, providing historical optimization references for adaptive refinement in future executions. A detailed explanation of each component is provided in Section~\ref{sec:methodology} }
\label{fig:static-dynamic}
\end{figure*}

\section{Methodology}
\label{sec:methodology}
This section details the Agentic Neural Network ($\mathcal{ANN}$) methodology, a multi-agent system framework designed to solve complex, multi-step computational tasks.  $\mathcal{ANN}$ is inspired by classic neural networks but replaces numerical weight optimizations with dynamic agent-based team selection and iterative textual refinement. Figure~\ref{fig:static-dynamic} shows the architecture of our $\mathcal{ANN}$.By structuring multi-agent collaboration hierarchically, $\mathcal{ANN}$ enables dynamic role assignment, adaptive aggregation, and data-driven coordination improvements through a forward-pass team selection process and a backward-pass optimization strategy. Importantly, the backward optimization is performed only during training on a task-family split to learn reusable agent-team pools and prompts, while inference on unseen tasks is strictly forward-only with a frozen architecture.

\subsection{Forward Dynamic Team Selection}
The $\mathcal{ANN}$ framework initiates task processing by decomposing the problem into structured subtasks. These subtasks are assigned across multiple layers, where each layer comprises a team of specialized agents working collaboratively on their designated subtask. Unlike static multi-agent workflows, $\mathcal{ANN}$ dynamically constructs these teams and their aggregation mechanisms based on task complexity. Two primary processes guide this phase: (1) defining the $\mathcal{ANN}$ structure and (2) selecting agentic teams that determine how agents collaborate.
\subsubsection{Structure of the Agentic Neural Network}

The architecture of $\mathcal{ANN}$ is inspired by neural networks, where each layer consists of nodes represented by agents. These agents are connected in a sequence that facilitates seamless information flow from one layer to the next, ensuring that outputs from a layer serve as structured inputs for the subsequent layer. This modular yet interconnected design enables efficient data processing, flexible task decomposition, and adaptive decision-making. Unlike static agent configurations, $\mathcal{ANN}$ dynamically refines its internal collaboration structure based on performance feedback, enhancing scalability and adaptability.
\subsubsection{Selection of Layer-wise Agentic Teams}
At each layer, $\mathcal{ANN}$ employs a mechanism to dynamically determine the most appropriate agentic teams, which dictates how multiple agents collaborate. This selection process considers the specific subtask requirements and complexity, ensuring that the most suitable collaborative strategy is applied to maximize performance. 

Let $\mathcal{F}_\ell$ be the set of candidate agent teams available for layer  $\ell$, $I_\ell$ the input to the layer, and $I$ the task-specific information. The dynamic team selection at each layer is determined by 
\vspace{-0.6em}
\begin{equation*}
    f_\ell = \text{DynamicRoutingSelect}(\mathcal{F}_\ell, \ell, I_\ell, I),
\end{equation*}
\vspace{-0.1em}
where $\text{DynamicRoutingSelect}$ denotes a process in which a team selector agent chooses the most appropriate agentic team from the candidate set based on task complexity.
The selection is informed by both subtask-specific information and a structural summary of each candidate team, which are provided to the team selector agent(See Appendix~\ref{sec:promptexample}.Prompt~\ref{prompt:6} for details). Here, $f_{\ell}$ represents the selected agentic team. Once an agentic team is selected, the layer processes input as: 
\vspace{-0.6em}
\begin{equation*}
    O_\ell = \text{ExecuteLayer}(\ell, f_\ell, I_\ell, I), 
\end{equation*}
\vspace{-0.1em}
where the selected agentic team $f_\ell$ is incorporated into the overall workflow and is responsible for executing the subtask assigned to layer~$\ell$. The resulting output $O_\ell$ is then propagated as the input to the next layer, i.e., $I_{\ell+1} = O_\ell$. This dynamic selection mechanism enables $\mathcal{ANN}$ to adapt to changing task conditions, thereby optimizing efficiency and accuracy in multi-agent collaboration.

\subsection{Backward Optimization}
During the training phase, upon completion of the forward execution on a training task, the system evaluates its performance. If the predefined performance thresholds are not met, $\mathcal{ANN}$ triggers a backward optimization phase to refine agentic collaboration team at both the global (system-wide) and local (layer-specific) levels. Textual gradients in $\mathcal{ANN}$ should be understood as structured, language-based optimization signals and suggestions rather than formal mathematical derivatives.
\subsubsection{Global Optimization}
Global optimization analyzes inter-layer coordination, refining interconnections and data flow to improve overall system performance. 
This process provides suggestions for updating the subtask descriptions of each layer and optimizing the configuration of agent teams at the inter-layer level, primarily identifying the modules that require structural improvement. 
As shown in Appendix~\ref{sec:promptexample}.Prompt~\ref{prompt:3} for details, it further optimizes information transfer across layers to ensure that local layer behaviors align with global objectives. 
Mathematically, the global gradient is computed as: 
\vspace{-0.6em}
\begin{equation*}
    \mathcal{G}_{\text{global}} = \text{ComputeGlobalGradient}(S, \tau), 
\end{equation*}
\vspace{-0.2em}
where $S$ represents the global workflow, and $\tau$ denotes the trajectory of execution, which includes agent interactions and input-output information transformations. 
The system structure is then updated accordingly: 
\vspace{-0.6em}
\begin{equation*}
    \mathcal{S}_{\text{global}} \gets \text{GlobalGradientUpdate}(\mathcal{G}_{\text{global}}, \tau). 
\end{equation*}
\vspace{-0.2em}
The resulting optimization suggestions are subsequently used to refine the structural composition of agent teams, enabling more coherent coordination across layers.

\subsubsection{Local Optimization}
While global optimization refines inter-layer interactions, local optimization focuses on improving the collaboration structure of agents within each layer, adjusting their nodes, edges, and prompts based on fine-grained performance feedback. 
The local gradient for each layer is computed as: 
\vspace{-0.6em}
\begin{align*}
\mathcal{G}_{\text{local},\ell}^{t} &= \beta \mathcal{G}_{\text{global}} + (1 - \beta) \\
&\quad \times \text{ComputeLocalGradient}(\ell, f_{\ell}, \tau), 
\end{align*} 
\vspace{-0.2em}
where $\beta$ is a weighting factor that balances the influence of global optimization and layer-specific gradients. 
At the $t$-th step, the agentic team is updated as: 
\vspace{-0.8em}
\begin{equation*}
    f_{\ell}^{t+1} = f_{\ell}^{t}  \times \mathcal{G}_{\text{local},\ell}^{t}, 
\end{equation*}
\vspace{-0.2em}
where the update is guided by the local gradient feedback, which provides suggestions to refine the structural configuration and prompt design of the current agentic team. 
The resulting optimized team $f_{\ell}^{t+1}$ integrates these improvements to enhance layer-level performance. Appendix~\ref{sec:pseudocode} provides pseudo-code and Appendix~\ref{sec:promptexample}.Prompt~\ref{prompt:4} provides prompts used to obtain textual local gradients.
\textbf{Momentum.} To improve stability, $\mathcal{ANN}$ employs momentum-based optimization, preventing sudden changes in agent parameters. The momentum-adjusted update rule is:
\vspace{-0.6em}
\begin{equation*}
    \mathcal{G}_{\text{local},\ell'}^{t} = \alpha \mathcal{G}_{\text{local},\ell}^{t} + (1 - \alpha) \mathcal{G}_{\text{local},\ell}^{t-1}, 
\end{equation*}
\vspace{-0.2em}
where $\alpha$ is the momentum coefficient, controlling how past updates influence the current optimization step. In essence, momentum provides the model with a memory of past optimization directions. 
As illustrated in Appendix~\ref{sec:promptexample}.Prompt~\ref{prompt:5}, it records how an agent team was previously adjusted (the previous adjustment direction) and combines this historical trajectory with the current feedback. By analyzing why the previous optimization failed to solve the present task, the framework refines the update direction, ensuring that new improvements build upon, rather than contradict, earlier progress. This helps the system maintain a clearer and stable sense of optimization direction across iterations.\\
\textbf{Format Validation.} It verifies that the updated structures conform to standard agent interaction formats, including proper edge definitions, variable references, and prompt templates, thereby maintaining consistency and reliability across the system.\\
\textbf{Performance Validation.} A validation agent evaluates the newly generated team against prior configurations, measuring structural similarity to prevent redundant designs and ensuring that each adjustment contributes positively to the overall functionality.



\section{Experiments}




\subsection{Experimental Settings}
\paragraph{Datasets} We evaluate our framework on seven benchmarks: HumanEval\cite{chen2021evaluatinglargelanguagemodels}, Creative Writing\cite{zhou2024symbolic}, MATH\cite{hendrycks2021measuringmathematicalproblemsolving}, DABench\cite{Hu2024InfiAgentDABenchEA}, MMLU\cite{hendrycks2021measuringmassivemultitasklanguage}, Natural Plan\cite{zheng2024naturalplanbenchmarkingllms} and AIME2024\&2025. HumanEval contains human-written coding problems and remains a standard benchmark for code generation. Creative Writing provides four-sentence prompts to craft a coherent story that ends with those sentences. MATH compiles challenging competition problems that demand multi-step symbolic reasoning across diverse fields. DABench covers data-analysis tasks such as feature engineering and statistics;  MMLU–Machine Learning is a subset from the Massive Multitask Language Understanding (MMLU) benchmark and offers multiple-choice questions on core ML concepts. Natural Plan focuses on scheduling tasks that require proposing feasible times under availability constraints in natural language.
Following prior work \cite{zhou2024symbolic,song2024adaptiveinconversationteambuilding,yuksekgonul2024textgrad,zhang2025swarmagenticfullyautomatedagentic}, we split the datasets into training and validation sets for each benchmark, ensuring direct comparability with their reported baselines. 


\paragraph{LLM Backbones}
To contain costs while maintaining strong performance, we unify the training process using the GPT-4o-mini model \cite{achiam2023gpt} for optimization. During validation, we evaluate datasets HumanEval, CW, MATH, DABench using three backbone variants: GPT-3.5-turbo, GPT-4o-mini, and GPT-4. Because neither \cite{zhou2024symbolic} nor \cite{song2024adaptiveinconversationteambuilding} report GPT-4o-mini results, our findings add a new dimension to the performance landscape, showing how a budget-friendly large language model can still match or surpass top-tier methods on standard tasks.
We evaluate datasets MMLU and Natural Plan using model GPT-4o and GPT-4o-0806 to follow the previous work . This setup enables us to demonstrate that our approach generalizes across different model capacities. We aim to demonstrate the flexibility and robustness of our framework in real-world various scenarios.

\paragraph{Baselines and Comparisons}
We compare \textbf{$\mathcal{ANN}$} (ours) against a broad range of representative baselines, including
GPTs~\cite{brown2020languagemodelsfewshotlearners,chen2021evaluatinglargelanguagemodels},
Agents~\cite{zhou2023agentsopensourceframeworkautonomous},
Agents w/ AutoPE~\cite{yang2024largelanguagemodelsoptimizers},
DSPy/ToT~\cite{khattab2023dspycompilingdeclarativelanguage},
Symbolic~\cite{zhou2024symbolic},
Vanilla LLM,
Meta-prompting~\cite{suzgun2024metapromptingenhancinglanguagemodels},
AutoAgents~\cite{chen2024autoagentsframeworkautomaticagent},
DyLAN~\cite{zhang2024dylan},
AgentVerse~\cite{chen2024agentverse},
AutoGen~\cite{wu2023autogen},
Captain Agent~\cite{song2024adaptiveinconversationteambuilding},
CoT~\cite{wei2023chainofthoughtpromptingelicitsreasoning},
TextGrad~\cite{yuksekgonul2024textgrad},
and ADAS~\cite{hu2025automated}.
Detailed descriptions of each baseline are provided in Appendix~\ref{sec:baseline_details}.

\subsection{Experimental Results}

\subsubsection{Main Results}

\begin{table}[t]
  \centering
  \renewcommand{\arraystretch}{1.5} 
  \footnotesize  
  \setlength{\tabcolsep}{3.5pt} 
  \begin{tabular}{l >{\centering\arraybackslash}p{2.3cm} >{\centering\arraybackslash}p{2.3cm}}
    \hline
    \rowcolor{gray!20}
    \textbf{Method} & \textbf{HumanEval} & \textbf{Creative Writing} \\
    \rowcolor{gray!10}
    & gpt-3.5/4o-mini/4 & gpt-3.5/4o-mini/4 \\
    \hline
    GPTs & 59.2 / - / 71.7 & 4.0 / - / 6.0 \\
    Agents & 59.5 / - / 85.0 & 4.2 / - / 6.0 \\
    Agents w/ AutoPE & 63.5 / - / 82.3 & 4.4 / - / 6.5 \\
    DSPy / ToT & 66.7 / - / 77.3 & 3.8 / - / 6.8 \\
    Symbolic & 64.5 / - / 85.8 & 6.9 / - / 7.4 \\
    
    \rowcolor{gray!10}
    \textbf{$\mathcal{ANN}$ (ours)} & \textbf{72.7} / 90.9 / \textbf{87.8} & \textbf{9.0} / 8.6 / \textbf{7.9} \\
    \hline
  \end{tabular}
  \caption{Comparison results on HumanEval and Creative Writing benchmarks. The best results in each category are marked in bold.}
  \label{table:humaneval_creative}
\end{table}

\begin{table}[t]
  \centering
  \renewcommand{\arraystretch}{1.5} 
  \footnotesize  
  \setlength{\tabcolsep}{3.5pt} 
  \begin{tabular}{>{\raggedright\arraybackslash}m{2.7cm} >{\centering\arraybackslash}p{1.8cm} >{\centering\arraybackslash}p{1.8cm}}
    \hline
    \rowcolor{gray!20}
    \textbf{Method} & \textbf{MATH} & \textbf{DABench} \\
    \hline
    Vanilla LLM & 51.53 &  6.61 \\
    Meta-prompting &68.88 &  39.69 \\
    AutoAgents &  56.12 &  57.98 \\
    DyLAN & 62.24 &  - \\
    AgentVerse &  69.38 &  - \\
    AutoGen & 74.49 &  82.88 \\
    Captain Agent &  77.55 &  88.32 \\
   
    \rowcolor{gray!10}
    $\mathcal{ANN}$ (gpt-4)  & \underline{80.0} & \textbf{90.2} \\
    \rowcolor{gray!10}
    $\mathcal{ANN}$ (gpt-3.5-turbo)        & 55.0&  76.0  \\
    \rowcolor{gray!10}
    $\mathcal{ANN}$ (gpt-4o-mini)         & \textbf{82.5} & \textbf{90.2}  \\
    \hline
  \end{tabular}
  \caption{Comparison results on the MATH and DABench datasets. The best results in each column are marked in bold, and the second-best results are underlined. All results without special annotation are based on GPT-4.}

  \label{table:math_dabench}
\end{table}

Table~\ref{table:humaneval_creative} compares our method with prior approaches on HumanEval and CW. Because \cite{zhou2024symbolic} provide baseline results only for GPT-3.5-turbo, hereafter referred to as GPT-3.5 and GPT-4, we supplement these with our own evaluations under GPT-4o-mini for a thorough comparison. We note the following key findings:
 On HumanEval, our $\mathcal{ANN}$ approach consistently surpasses all baselines. We achieve \textbf{72.7}\% and \textbf{87.8}\% for GPT-3.5 and GPT-4, respectively, outperforming the best baseline by a clear margin. Notably, even our GPT-4o-mini results(\textbf{90.9}\%)  show competitive or superior performance despite GPT-4o-mini being a lower-cost model.
For open-ended text generation tasks in CW, our method scores \textbf{9.0}/\textbf{7.9} on GPT-3.5/GPT-4. We attribute this to $\mathcal{ANN}$’s structured layer-wise approach, which fosters creative synergy among specialized agents while maintaining logical consistency in narrative structure.

In Table~\ref{table:math_dabench}, we contrast our method with baseline results from \cite{song2024adaptiveinconversationteambuilding} on MATH and DABench. Notably, \cite{song2024adaptiveinconversationteambuilding} report results using GPT-4 but omit GPT-3.5 and GPT-4o-mini. On MATH, We record 55.0, \textbf{82.5}, and 80.0 across GPT-3.5, 4o-mini, and GPT-4. Despite using GPT-4o-mini in training, our method exhibits strong generalization to both GPT-3.5 and GPT-4. On GPT-4, our \textbf{80.0}\% accuracy significantly outperforms Captain Agent (77.55\%) and AutoGen (74.49\%). On DABench, which focuses on data-analysis tasks, our method ($\mathcal{ANN}$) attains 76.0\%, \textbf{90.2\%}, and 90.2\% on GPT-3.5, GPT-4o-mini, and GPT-4, respectively, consistently outperforming prior baselines. We observe that GPT-4o-mini again surprisingly yields top-tier results, indicating that data-centric tasks can benefit from well-structured agent orchestration without always requiring the largest language models.

Following \cite{zhang2025swarmagenticfullyautomatedagentic}, it employs distinct models
for optimization and execution. Specifically, we
use GPT-4o-mini for training-phase with optimization, and select GPT-4o for validation-phase only with execution.
We contrast our method with baseline results from \cite{yuksekgonul2024textgrad,zhang2025swarmagenticfullyautomatedagentic} on the MMLU-ML and Natural Plan (see Table~\ref{table:mmlu}. For MMLU, our method achieves \textbf{89.2}\% accuracy, outperforming CoT \cite{wei2023chainofthoughtpromptingelicitsreasoning} (85.7\%) and TextGrad \cite{yuksekgonul2024textgrad} (88.4\%). For the Natural Plan benchmark, our method achieves accuracies of \textbf{7.9/55.0/73.0}, outperforming CoT (1.0/50.0/60.0) and ADAS \cite{hu2025automated} (3.1/43.0/66.0). This result demonstrates the advantage of our layerwise optimization approach in highly structured reasoning settings.

\begin{table}[t]
  \centering
  \renewcommand{\arraystretch}{1.25}
  \setlength{\tabcolsep}{5pt} 
  \footnotesize
  \begin{tabular}{p{1cm} c ccc}
    \rowcolor{gray!10}
    \textbf{Method} & \textbf{MMLU} & \multicolumn{3}{c}{\textbf{Natural Plan}} \\
    \rowcolor{gray!10}
    & \textbf{ML} & \makecell{\textbf{Trip}\\\textbf{Planning}}  & \makecell{\textbf{Meeting}\\\textbf{Planning}}  & \makecell{\textbf{Calendar}\\\textbf{Scheduling}} \\
    \cmidrule(lr){3-5}
    CoT & 85.7 & 1.0 & 50.0 & 60.0 \\
    ADAS & -- & 3.1 & 43.0 & 66.0 \\
    TextGrad & 88.4 & -- & -- & -- \\
    \rowcolor{gray!10}
    \textbf{$\mathcal{ANN}$} & \textbf{89.2} & \textbf{7.9} & \textbf{55.0} & \textbf{73.0} \\
    \hline
  \end{tabular}
  \caption{Accuracy on MMLU-ML \protect\footnotemark~and the Natural Plan benchmark, which consists of three task domains: TP (Trip Planning), MP (Meeting Planning), and CS (Calendar Scheduling).}
  \label{table:mmlu}
\end{table}
\footnotetext{We report results on the MMLU-ML subset for fair comparison, as TextGrad provides separate results on this subset.}

\subsubsection{Robustness to Backbone Variation}
 To address concerns regarding our use of a single backbone during training, we conducted an additional experiment using GPT-3.5-turbo as the training model while retaining GPT-3.5-turbo, GPT-4o-mini, and GPT-4 as evaluation backbones. Results across HumanEval, CW, Math, and DABench benchmarks (see Table~\ref{table:appendix_backbone_generalization}) show that $\mathcal{ANN}$ achieves strong generalization even when trained on GPT-3.5-turbo, a smaller-capacity model.
 This suggests that the agentic orchestration and textual backpropagation mechanisms in $\mathcal{ANN}$ are robust to changes in underlying language model capacity.
Experiments demonstrate that the multi-agent architecture discovered by our $\mathcal{ANN}$ framework, even when using the weaker GPT-4o-mini, can generalize effectively to more powerful LLMs, achieving superior performance. Additionally, our results highlight GPT-4o-mini as a cost-effective yet high-performing alternative, reinforcing $\mathcal{ANN}$’s robustness across different model scales.

\begin{table*}[t]
  \centering
  \renewcommand{\arraystretch}{1.5}
  \footnotesize
  \setlength{\tabcolsep}{5pt}
  \rowcolors{3}{}{gray!5}

  \begin{tabular}{
    >{\raggedright\arraybackslash}m{1.8cm} 
    >{\centering\arraybackslash}m{2.5cm} 
    >{\centering\arraybackslash}m{2.5cm} 
    >{\centering\arraybackslash}m{2.5cm} 
    >{\centering\arraybackslash}m{2.5cm}
    >{\centering\arraybackslash}m{2.2cm}
  }
    \rowcolor{gray!20}
    \textbf{Train / Eval} & \textbf{HumanEval} & \textbf{Creative Writing} & \textbf{MATH} & \textbf{DABench} & \textit{Total Train Cost} \\
    \rowcolor{gray!10}
     \textbf{ Backbones} & GPT-3.5/4o-mini/4 & GPT-3.5/4o-mini/4 & GPT-3.5/4o-mini/4 & GPT-3.5/4o-mini/4 & (in USD) \\
    \hline
    GPT-3.5 & 73.7 / 72.7 / 86.3 & 8.9 / 8.5 / 8.1 & 53.5 / 80.0 / 80.0 & 70.6 / 88.0 / 90.2 & $\approx$\$122.30 \\
    GPT-4o-mini  & 72.7 / 90.9 / 87.8 & 9.0 / 8.6 / 7.9 & 55.0 / 82.5 / 80.0 & 76.0 / 90.2 / 90.2 & $\approx$\$73.40 \\
    \hline
  \end{tabular}

  \caption{
    \textbf{Evaluation results across four benchmarks (HumanEval, CW, Math, and DABench)} with two different training backbones (GPT-3.5 vs GPT-4o-mini), evaluated across GPT-3.5, GPT-4o-mini, and GPT-4. Training costs are estimated based on approximately 244.6M input tokens.
  }
  \label{table:appendix_backbone_generalization}
\end{table*}

\subsubsection{Ablation Studies}
\label{sec:ablation}
\begin{figure*}[t]
  \includegraphics[width=0.25\linewidth,height=0.20\linewidth]{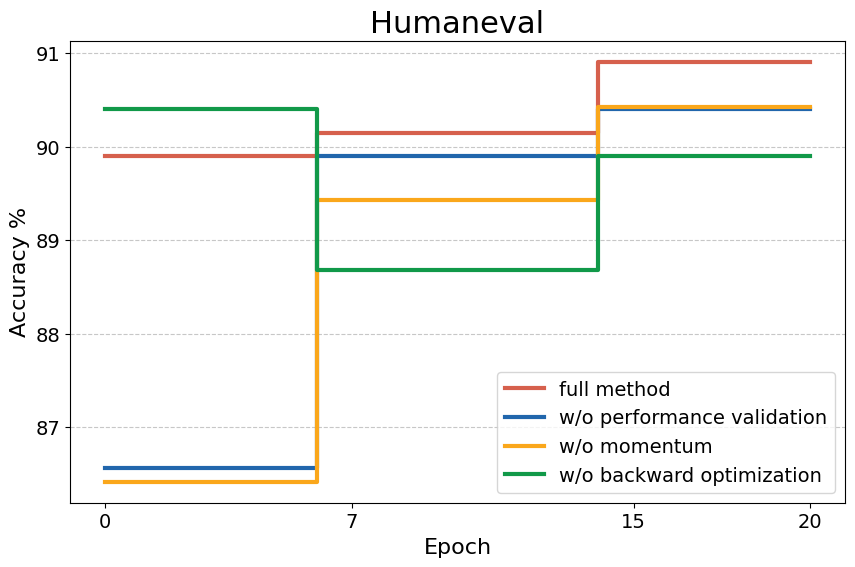}\hfill
  \includegraphics[width=0.25\linewidth,height=0.20\linewidth]{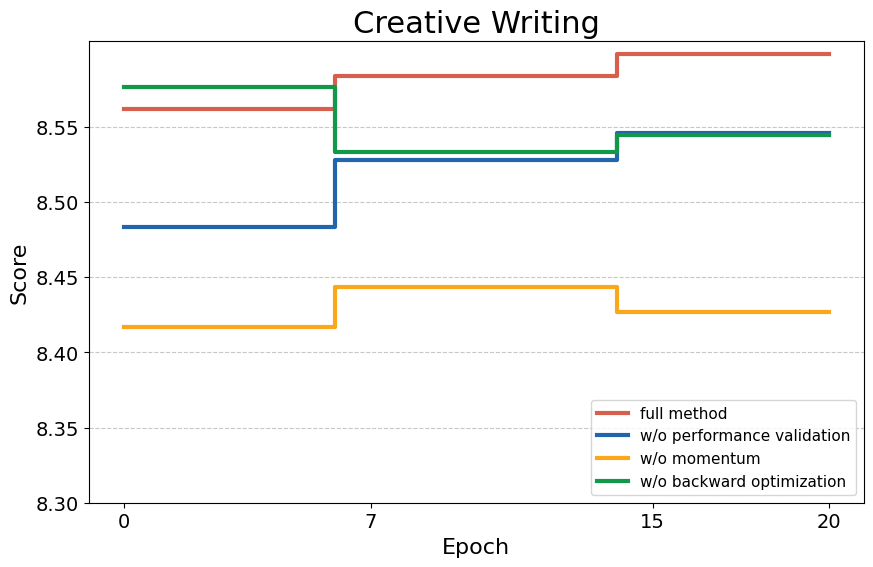}\hfill
  \includegraphics[width=0.25\linewidth,height=0.20\linewidth]{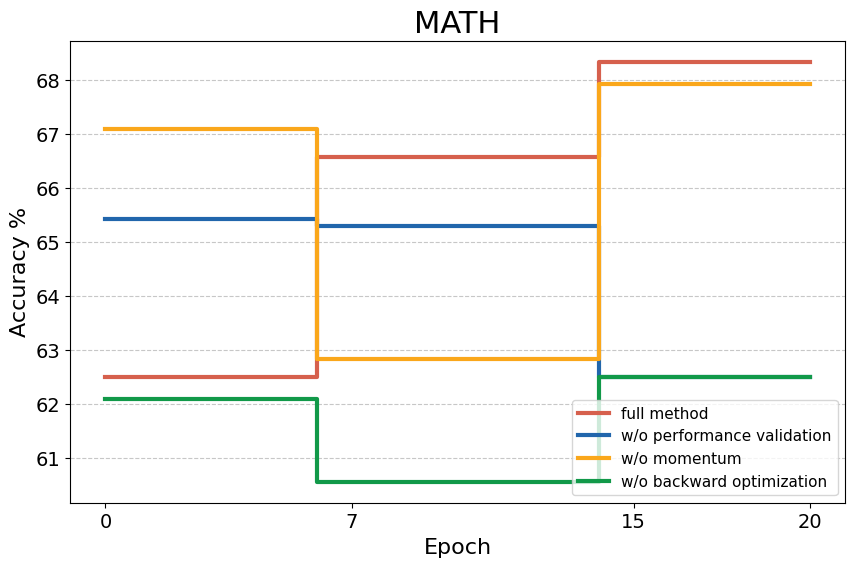}\hfill
  \includegraphics[width=0.25\linewidth,height=0.20\linewidth]{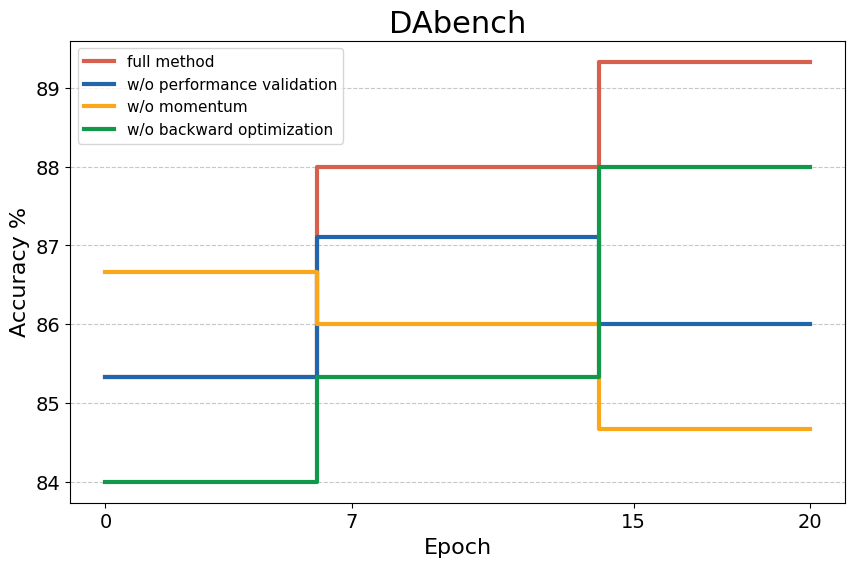}
\caption{
Ablation results on HumanEval, CW, MATH, and DABench using the GPT-4o-mini model for both training and validation. We compare the full $\mathcal{ANN}$ framework (red curve) against three ablated variants: 
w/o Validation Performance (blue curve), 
w/o Momentum (purple curve), and 
w/o Backward Optimization (green curve). 
Each curve shows average validation accuracy (or equivalent score) over three runs. 
The full $\mathcal{ANN}$ consistently outperforms all ablations, confirming the necessity of each component.
}
\label{fig:ablation_plot}
\end{figure*}


We conduct a unified ablation study using only GPT-4o-mini to further investigate the design choices in our $\mathcal{ANN}$ framework. Specifically, we compare four variants: (1) \textbf{Full $\mathcal{ANN}$:} Our complete approach with momentum-based optimization, validation-based performance checks, and backward optimization. (2) \textbf{w/o Momentum:} Disables the momentum technique in textual gradient refinement. (3)\textbf{w/o Validation Performance:} Skips the validation-based filtering stage when selecting improved prompts and agent roles. (4)\textbf{w/o Backward Optimization:} Does not use the backward pass to refine prompts, i.e., textual gradients for \textit{error signals} are omitted.
\paragraph{Training Procedure.}
All four variants are trained for 20 epochs on each dataset (HumanEval, CW, MATH, DABench) using the training splits described above. To mitigate the randomness inherent in LLM sampling, we repeat each condition \emph{three times} and report the \emph{average} results on the validation set at regular epoch intervals.  
\paragraph{Results and Analysis.}
Figure~\ref{fig:ablation_plot} illustrates the validation accuracy (or relevant score) as a function of training epoch. We observe a consistent upward trend across all four datasets, with the full $\mathcal{ANN}$ approach converging to the highest performance. Detailed findings indicate that the \textbf{impact of momentum} is substantial: removing momentum (w/o Momentum) leads to the largest performance drop on HumanEval, suggesting that gradual accumulation of textual gradient signals is crucial for code-generation tasks that require precise correctness. \textbf{Validation-based checks} also play an important role—omitting validation performance filtering can cause more erratic updates, particularly evident in MATH, where narrative consistency can degrade if suboptimal agent prompts are accepted too frequently. Finally, \textbf{backward optimization} proves essential: without the backward pass, we lose a key mechanism for refining agentic team. Overall, our ablation highlights that each component contributes significantly to performance, and combining them yields the most reliable and robust improvements.

\subsubsection{Cross-Dataset Transfer on Mathematical Benchmarks}
\begin{figure}[t]
  \centering
  \includegraphics[width=0.9\linewidth]{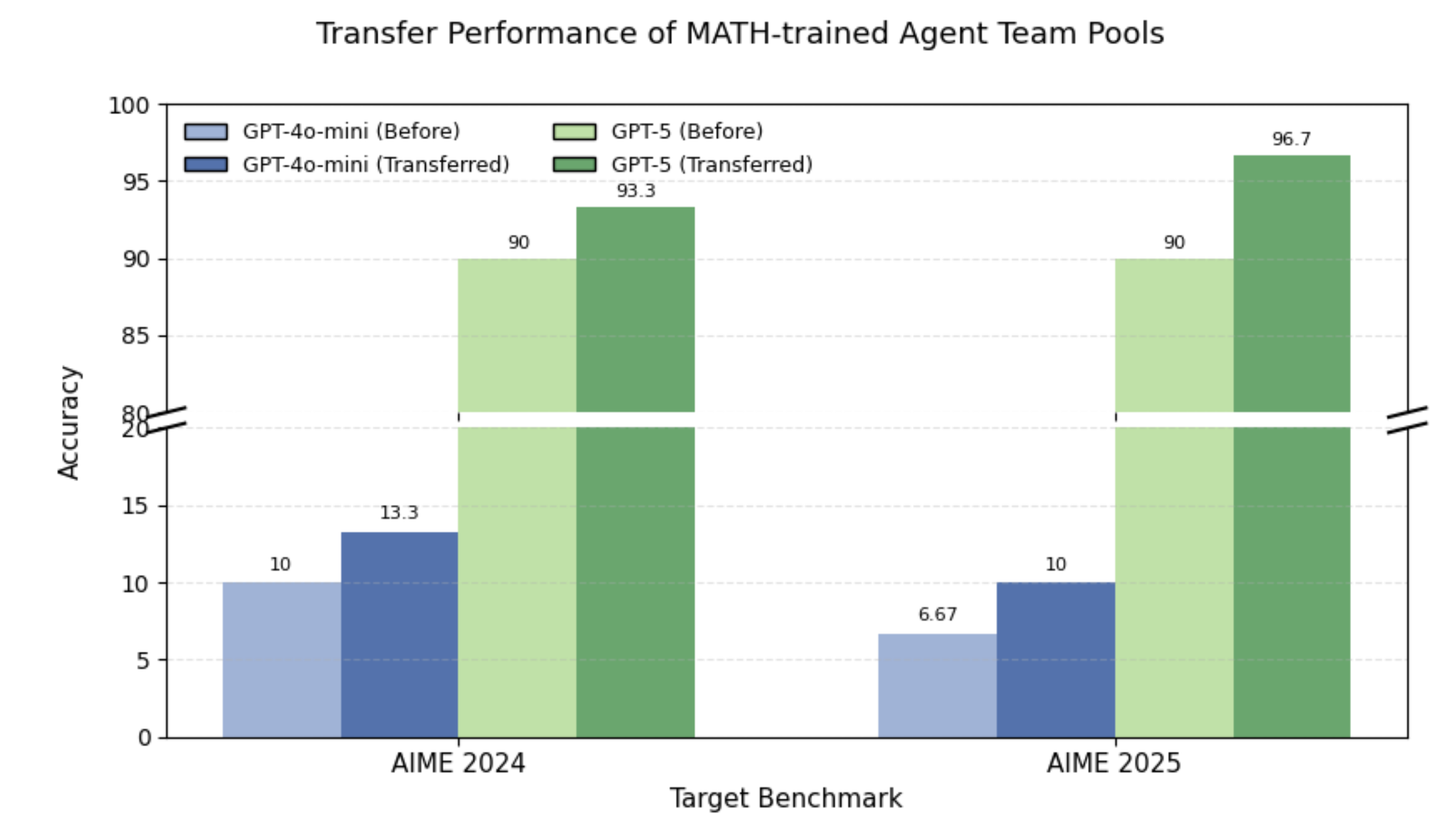}
  \caption{
 Cross-dataset transfer performance of MATH-trained agent team pools on AIME 2024 and AIME 2025 without further backward optimization.}
  \label{fig:math_transfer}
\end{figure}
To further evaluate whether $\mathcal{ANN}$ requires re-optimization for each new task or can generalize across related task families, we conduct a cross-dataset transfer study on mathematical reasoning benchmarks.
We first optimize the agent team pools on the MATH training split, and then directly apply the learned candidate pools to AIME 2024 and AIME 2025 without any further backward optimization. Figure~\ref{fig:math_transfer} shows that agent team pools optimized on MATH consistently improve performance on both AIME benchmarks after training.
Notably, the transferred pools yield clear gains over the unoptimized (before) configurations under both GPT-4o-mini and GPT-5 backbones.
This demonstrates that $\mathcal{ANN}$ learns reusable agentic structures that transfer across related mathematical problem distributions, rather than overfitting to a single dataset or requiring per-task re-optimization.

\section{Conclusion}
Our experimental results establish that $\mathcal{ANN}$ achieves high accuracy and adaptability across tasks ranging from code generation to creative writing, surpassing traditional static configurations. Through a dynamic formation of agent teams and a two-phase optimization pipeline, the framework delivers robust performance rooted in neural network design principles. These findings underscore the potential of $\mathcal{ANN}$ as an efficient solution for orchestrating complex multi-agent workflows. Detailed ablation studies highlight the significance of each component. Ultimately, this integrated agentic paradigm enables self-evolving multi-agent systems through coordinated optimization.

\section*{Limitations}

Despite its advantages, the Agentic Neural Network framework has limitations. While $\mathcal{ANN}$ significantly reduces manual effort compared to prior multi-agent systems, it still requires a small amount of human initialization in the form of lightweight agent team structures and prompts. This design choice prioritizes reliability and controllability, but prevents full end-to-end automation.
Future work will explore meta-prompt learning and dynamic role adjustment to further minimize human intervention while preserving stability and performance.


\section*{Acknowledgements}

The authors gratefully acknowledge the scientific support and HPC resources provided by the Karlsruhe Institute of Technology National High
Performance Computing Center (NHR@KIT) under NHR projects 25312 and 22560.

\bibliography{custom}

\newpage
\clearpage
\appendix
\onecolumn
\tableofcontents

\newpage

\section{Comparison and baseline}
\label{sec:comparision}


\subsection{Comparision}
\subsubsection{Framework-Level Comparison}
With the rapid advancement and widespread adoption of deep learning techniques \cite{liu2022frfoldedrationalizationunified,liu2023mgrmultigeneratorbasedrationalization,lu2019singleimagesuperresolution,Lu_2022,Lu_2023,tian2023heterogeneous,tian2024graph,NOSMOG_ICLR,FGGP_AAAI24,liu2025gated,xiaofedloge,NEURIPS2023_f4b8ddb9,wang2025ascd,jiang2025minedprobingupdatingmultimodal,jiang2025koreenhancingknowledgeinjection,zhang2023spot,peng2025visualinputcompressedvisual,ReTrack,HABIT,ENCODER,FineCIR,OFFSET,HUD,INTENT,REFINE}, large language models \cite{bi2024visual,bi2025cot,bi2025prism,du2025graftllm,du2025npspruning,wang2025evaluating,huang2025loongsynthesizelongchainofthoughts,yang2026alignsaeconceptalignedsparseautoencoders,wan2025magicwordssharpnessawareprompt,tian2025reinforcementmidtraining,li2026graphsubstratedatamodalities,wan2025hyperion,wang2025geovistawebaugmentedagenticvisual,wang-etal-2025-visuothink} have emerged as a transformative force across diverse domains \cite{Chen_2025_CVPR,rong2025backdoor,zhang2023spot,chen2025does,zhao2024large,yu2025prnet,huang2024t2i,zeng2024mitigating,xu-etal-2024-llm,lu2024mace,lu2023tf,rong2025can,liu2022frfoldedrationalizationunified,wang2025geometryzeroimprovinggeometrysolving,wang2024rescuerankingllmresponses,wang2025visuothinkempoweringlvlmreasoning,wang-etal-2024-conu,wang-etal-2025-sconu,wang2025coin,wang2025lec,wang2025word,yuan2025kardia,yuan2026query,yuan2026decoder,huang2025competition,jiang2025danmakutppbench,jiang2025satiredecoder,du2024pcb,du2025graftllm,du2025nps,zhao2025autoreproduce,zhao2025chartcoder,zhao2025vincicoder,zhao2026tinychemvl,yang2026omnidiagramadvancingunifieddiagram,zhang1,zhang2,zhang3,zhang4,zhang5,zhang6,zhang7,zhang8,zhang9}. Their ability to understand, generate, and reason over natural language has enabled a new generation of intelligent systems, particularly in the orchestration and coordination of multi-agent frameworks. As these models continue to evolve, numerous architectures have been proposed to harness their capabilities in increasingly sophisticated and dynamic environments.

To situate $\mathcal{ANN}$ in the rapidly evolving ecosystem of multi-agent orchestration, we benchmark it against nine representative frameworks drawn from recent literature—\emph{Symbolic}~\cite{zhou2024symbolic}, \emph{AutoGen}~\cite{wu2023autogen}, \emph{InfiAgent-DAbench}~\cite{Hu2024InfiAgentDABenchEA}, \emph{MetaGPT}~\cite{Hong2023MetaGPTMP}, \emph{DyLan}~\cite{zhang2024dylan}, \emph{Adaptive Team}~\cite{song2024adaptiveinconversationteambuilding}, \emph{Chain-of-Thought}~\cite{wei2023chainofthoughtpromptingelicitsreasoning}, \emph{GPTSwarm}~\cite{Zhuge2024LanguageAA}, and \emph{Aflow}~\cite{zhang2024autoflow}.  
Collectively, these baselines cover symbolic planning, agentic workflow coordination, dynamic team formation, and optimisation-driven routines, thus furnishing a balanced backdrop for assessing architectural and functional advances.

Table~\ref{tab:framework_comparison} distils the comparison along seven orthogonal dimensions: (i) \emph{layerwise decomposition}, (ii) \emph{back-propagated optimisation}, (iii) \emph{momentum-based adjustment}, (iv) \emph{global optimisation scope}, (v) \emph{local-only optimisation}, (vi) \emph{dynamic team selection}, and (vii) \emph{task-specific training requirements}.  
A check mark (\cmark) indicates native support; a cross (\xmark) denotes absence.  
As the table shows, $\mathcal{ANN}$ is the only framework that provides \emph{full} coverage across all criteria—combining layerwise granularity with momentum-augmented backward optimisation, unifying global and local objectives, and eliminating the need for costly task-specific fine-tuning through on-the-fly team selection.

\begin{table*}[t]
  \centering
  \renewcommand{\arraystretch}{1.5}
  \small
  \setlength{\tabcolsep}{5pt}
  \begin{tabular}{
    l
    >{\centering\arraybackslash}p{1.2cm}
    >{\centering\arraybackslash}p{1.2cm}
    >{\centering\arraybackslash}p{1.4cm}
    >{\centering\arraybackslash}p{1.2cm}
    >{\centering\arraybackslash}p{1.2cm}
    >{\centering\arraybackslash}p{1.2cm}
    >{\centering\arraybackslash}p{1.2cm}
  }
    \toprule
    \rowcolor{gray!10}
    \textbf{Framework} & \textbf{Layerwise} & \textbf{Backward Optimization} & \textbf{Momentum Adjustment} & \textbf{Global Optimization} & \textbf{Local Optimization} & \textbf{Dynamic
Teaming} & \textbf{Training Requirement} \\
    \hline
    \midrule
    Symbolic \cite{zhou2024symbolic})                & \xmark & \cmark & \xmark & \cmark & \cmark & \xmark & \cmark \\
    AutoGen \cite{wu2023autogen}                  & \xmark & \xmark & \xmark & \cmark & \cmark & \cmark & \xmark \\
    InfiAgent-DAbench \cite{Hu2024InfiAgentDABenchEA}               & \xmark & \xmark & \xmark & \cmark & \xmark & \cmark & \xmark \\
    MetaGPT \cite{Hong2023MetaGPTMP}                  & \xmark & \xmark & \xmark & \xmark & \cmark & \cmark & \xmark \\
    DyLan \cite{zhang2024dylan}                    & \xmark & \cmark & \xmark & \cmark & \cmark & \cmark & \cmark \\
    Adaptive Team \cite{song2024adaptiveinconversationteambuilding}                        & \xmark & \xmark & \cmark & \xmark & \cmark & \cmark & \xmark \\
    Chain-of-Thought \cite{wei2023chainofthoughtpromptingelicitsreasoning}                       & \xmark & \xmark & \xmark & \xmark & \cmark & \xmark & \xmark \\
    GPTSwarm \cite{Zhuge2024LanguageAA}                     & \xmark & \xmark & \cmark & \cmark & \cmark & \cmark & \cmark \\
    Aflow \cite{zhang2024autoflow}                     & \xmark & \xmark & \xmark & \cmark & \xmark & \cmark & \xmark \\
    \rowcolor{LightGray}
    \hline
    \textbf{$\mathcal{ANN}$ (Ours)}       & \cmark & \cmark & \cmark & \cmark & \cmark & \cmark & \cmark \\
    \bottomrule
  \end{tabular}
  \caption{
    Framework-level comparison across layerwise design, optimization strategies (backward, momentum, global/local), dynamic team composition, and training requirements. 
    \cmark/\xmark\ indicate support.
  }
  \label{tab:framework_comparison}
\end{table*}

\subsubsection{Trainable Architectures versus Test-Time Reasoning}

Our framework is inspired by planning- and reflection-style reasoning methods such as Chain-of-Thought (CoT)~\cite{wei2023chainofthoughtpromptingelicitsreasoning}, Tree-of-Thought (ToT)~\cite{khattab2023dspycompilingdeclarativelanguage}, Graph-of-Thought (GoT)~\cite{Besta_2024}, and Reflexion, which demonstrate that explicitly structuring intermediate reasoning steps, search trees, or reflection loops can significantly improve large language model performance on complex tasks.
However, $\mathcal{ANN}$ differs fundamentally from these methods in formulation, optimization, and reuse of multi-agent workflows.

\paragraph{Test-time reasoning vs.\ training-time optimization.}
Planning and reflection methods operate purely at \emph{test time}.
For each new input instance, they dynamically expand reasoning chains, trees, or reflection trajectories and discard them after inference.
They do not maintain a persistent architecture, nor do they perform training-phase optimization over agent structures.
As a result, performance is scaled primarily by increasing test-time computation rather than by learning reusable workflow structures.

In contrast, $\mathcal{ANN}$ treats a multi-agent workflow as a \emph{trainable, layered architecture}.
The framework defines a fixed number of layers, each maintaining a candidate pool of structured agent teams (blocks composed of nodes, edges, and prompts).
These candidate pools are optimized during a lightweight training phase using a small training split, and the resulting architecture is reused for all future instances at test time without further search or expansion.
This decouples architectural learning from test-time inference and positions $\mathcal{ANN}$ as a data-efficient training procedure rather than a test-time scaling heuristic.

\paragraph{Local reasoning updates vs.\ workflow-level optimization.}
Reflection-based methods typically revise a single reasoning trace, prompt, or trajectory in isolation.
In contrast, optimization in $\mathcal{ANN}$ is performed through a forward--backward loop over the \emph{entire layered workflow}.
During the backward phase, textual back-propagation decomposes feedback into a global component, which identifies structurally problematic layers and blocks, and local components, which refine nodes, edges, and prompts within selected blocks.
These textual gradients are accumulated with momentum, providing a notion of optimization trajectory across iterations and enabling stable architectural refinement.

\paragraph{Implicit reasoning structures vs.\ explicit agentic architectures.}
In $\mathcal{ANN}$, agent team structures are treated as discrete parameters subject to optimization.
At inference time, a team-selector agent performs dynamic routing by selecting the most suitable block from each layer’s candidate pool based on the current subtask and execution trajectory.
Importantly, the backward phase does not blindly accept LLM-generated updates: every proposed block must pass format, structure, and performance validation before entering the candidate pool.
This yields a neural-network-like training loop that searches over a discrete space of multi-agent architectures, as summarized in Table~\ref{tab:framework_comparison}.

Taken together, while $\mathcal{ANN}$ draws conceptual inspiration from planning and reflection methods, it differs fundamentally in scope and mechanism.
Rather than expanding reasoning at test time, our framework learns a reusable, layered multi-agent architecture through training-phase optimization, enabling stable generalization across tasks and instances.

\subsection{Baseline Details}
\label{sec:baseline_details}

We compare \textbf{$\mathcal{ANN}$} (ours) with the following baseline approaches:

\noindent
\textbf{GPTs}~\cite{brown2020languagemodelsfewshotlearners,chen2021evaluatinglargelanguagemodels}: 
A direct usage of GPT-based models with carefully designed prompts.

\noindent
\textbf{Agents}~\cite{zhou2023agentsopensourceframeworkautonomous}: 
A language-agent method that organizes multi-step reasoning and tool usage through a pipeline of prompts.

\noindent
\textbf{Agents w/ AutoPE}~\cite{yang2024largelanguagemodelsoptimizers}: 
A variant wherein each prompt node is optimized by an LLM, but without full language gradient backpropagation.

\noindent
\textbf{DSPy / ToT}~\cite{khattab2023dspycompilingdeclarativelanguage}: 
A pipeline optimization framework that performs search-based tuning of prompt components, applicable mostly to tasks with a tractable evaluation function.

\noindent
\textbf{Symbolic}~\cite{zhou2024symbolic}: 
An agent-based system employing symbolic learning methods for dynamic prompt improvements.

\noindent
\textbf{Vanilla LLM}: 
A single-turn GPT-based approach without agent collaboration.

\noindent
\textbf{Meta-prompting}~\cite{suzgun2024metapromptingenhancinglanguagemodels}: 
An adaptive prompting strategy that attempts to generate meta-level instructions for new tasks.

\noindent
\textbf{AutoAgents}~\cite{chen2024autoagentsframeworkautomaticagent}: 
An automated agent system that attempts to orchestrate multi-agent interactions but can be unstable in large-scale settings.

\noindent
\textbf{DyLAN}~\cite{zhang2024dylan}: 
A dynamic language-agent approach that decomposes tasks with feedback loops.

\noindent
\textbf{AgentVerse}~\cite{chen2024agentverse}: 
A multi-agent platform emphasizing flexible agent composition.

\noindent
\textbf{AutoGen}~\cite{wu2023autogen}: 
A system featuring an ``Assistant + Executor'' design for multi-step problem-solving.

\noindent
\textbf{Captain Agent}~\cite{song2024adaptiveinconversationteambuilding}: 
An adaptive team-building agent framework that spawns specialized sub-agents based on task progress.

\noindent
\textbf{CoT (Chain-of-Thought)}~\cite{wei2023chainofthoughtpromptingelicitsreasoning}: 
A prompting strategy that encourages intermediate reasoning steps, often used to enhance zero-shot performance on complex QA tasks.

\noindent
\textbf{TextGrad}~\cite{yuksekgonul2024textgrad}: 
A framework that performs solution-level optimization using textual gradients.

\noindent
\textbf{ADAS (Automated Design of Agentic Systems)}~\cite{hu2025automated}: 
A framework in which a meta-agent autonomously generates agentic architectures for multi-agent system design.

\section{Implementation}

\subsection{Pseudo Code}
\label{sec:pseudocode}
This section provides pseudocode for the system’s overall architecture and the local gradient optimization process. Algorithm~\ref{algo:1} outlines how the network leverages a dynamic routing mechanism alongside an agentic neural network structure, 
integrating both global optimization and layerwise optimization. Dynamic routing selects the most suitable path for a given task, thereby enhancing overall system performance and stability. Global optimization steers the entire network toward optimal solutions, while layerwise optimization fine-tunes each layer for improved learning efficiency and reliability.
Algorithm~\ref{algo:2} focuses on local optimization within each specialized layer. By applying localized gradient updates, each module can concentrate on its respective sub-task. Such targeted adjustments accelerate convergence, improve learning efficiency, and, in conjunction with the global optimization strategy, enhance the system’s overall performance.

\begin{figure*}[!t]
\refstepcounter{algorithm}  
\label{algo:1} 
\begin{tcolorbox}[algobox, title={Algorithm 1: Agentic Neural Network with Dynamic Routing and Adaptive Optimization}]
\begin{algorithmic}[1]
\Require $I$: dataset input; $L$: layers in the workflow; $F_\ell$: set of possible aggregation functions for each layer $\ell$; $S$: workflow updation for optimization
\Ensure Updated structure and prompts for the agentic neural network
\State $\text{Traj} \gets []$ \Comment{Initialize Trajectory}
\State $\text{$I_\ell$} \gets I$ \Comment{Initialize input of first layer}
\State \textbf{Forward Pass with Dynamic Routing and Aggregation}
\For{each layer $\ell$ in $L$}
    \State $f_\ell \gets \text{DynamicRoutingSelect}(F_\ell, \ell, I_\ell, I)$
    \State $O_\ell \gets \text{ExecuteLayer}(\ell, f_\ell, I_\ell, I)$
    \State Append $(\ell, f_\ell, I_\ell, O_\ell)$ to $\text{Traj}$
    \State $I_{\ell+1} \gets O_\ell$
\EndFor
\State \textbf{Backpropagation:}
\State \textbf{Global Optimization}
\State $\mathcal{G}_{\text{global}} \gets \text{ComputeGlobalGradient}(S, \text{Traj})$
\State $\mathcal{S}_{\text{global}} \gets \text{GlobalGradientUpdate}(\mathcal{G}_{\text{global}}, \text{Traj})$
\State \textbf{Layerwise Optimization}
\For{each layer $\ell$ in reverse($L$)}
    \State $\mathcal{G}_{local,\ell}^{t} \gets \text{ComputeLocalGradient}(\ell, f_\ell, \text{Traj}, \mathcal{L}_{\text{global}})$
    \If{momentum\_needed}
        \State $\mathcal{S}_{\text{local}} \gets \text{LocalGradientUpdate}(\ell,f_\ell, \mathcal{G}_{local,\ell}^{t}, \mathcal{S}_{\text{global}})$
    \Else
        \State $\mathcal{G}_{local,\ell'}^{t} \gets \text{ApplyMomentum}(\ell, \text{Traj}, \mathcal{G}_{local,\ell}^{t}, \mathcal{G}_{local,\ell}^{t-1})$
        \State $\mathcal{S}_{\text{local}} \gets \text{LocalGradientUpdate}(\ell,f_\ell, \mathcal{G}_{local,\ell'}^{t}, \mathcal{S}_{\text{global}})$
    \EndIf
\EndFor
\State \Return $(F_\ell, \text{Traj})$
\end{algorithmic}
\end{tcolorbox}
\end{figure*}

\begin{figure*}[!t]
\refstepcounter{algorithm}  
\label{algo:2} 
\begin{tcolorbox}[algobox,title={Algorithm 2: LocalGradientUpdate}]
\begin{algorithmic}[1]
\Require $\ell$: current layer; $f_\ell$: selected aggregation function; $\text{Traj}$: trajectory of execution; $\mathcal{G}_{\text{global}}$: global gradient; $\mathcal{S}_{\text{global}}$: current global structure; $F_\ell$: set of possible aggregation functions for each layer $\ell$
\Ensure Updated global structure $\mathcal{S}_{\text{global}}$ and valid aggregation function $f_\ell$

\State $\mathcal{G}_{local,\ell} \gets \text{ComputeLocalGradient}(\ell, f_\ell, \text{Traj}, \mathcal{G}_{\text{global}})$
\Comment{Compute local gradient in layer $\ell$}
\State $\mathcal{S}_{\text{local}} \gets \text{LocalGradientUpdate}(\ell,f_\ell, \mathcal{G}_{local,\ell}, \mathcal{S}_{\text{global}})$:
\Comment{$\mathcal{S}_{\text{local}}$: Update layer-wise workflow}
\For{$k \gets 1$ to 3} \Comment{Attempt up to 3 updates}
    \State $f'_{\ell} \gets \text{LocalGradientUpdate}(\ell, f_\ell, \mathcal{G}_{local,\ell}, \mathcal{S}_{\text{global}})$
    \State \textbf{ValidateUpdate} ($f'_\ell$): \Comment{If update passes validation}
        \State \textbf{Node Validation:}
        \If{\text{VariableSourcesValid}($f'_{\ell}$) \& \text{FormatValid}($f'_{\ell}$)}
            \State \textbf{Edge Validation:}
            \If{\text{AllNodesHaveEdges}($f'_{\ell}$)}
                \State \textbf{Structure Validation:}
                \If{\text{StructureNotUnique}($f'_{\ell}$)}
                    \If{\text{ValidatePerformance}($f'_\ell, f_\ell$)}
                        \State Append $f'_{\ell}$ to $F_{\ell}$ \Comment add new agg func $f'_{\ell}$ into $F_\ell$
                        \State \textbf{break} \Comment{Exit update loop on success}
                    \EndIf
                \EndIf
            \EndIf
        \EndIf
\EndFor

\State \Return $\mathcal{S}_{\text{global}}$
\end{algorithmic}
\end{tcolorbox}
\end{figure*}

\subsection{Initialization and Deployment}

Although $\mathcal{ANN}$ supports dynamic structural evolution through textual backpropagation, the amount of manual initialization required in our implementation is intentionally kept minimal.
Our design follows the same practical assumption adopted by prior multi-agent frameworks, namely that a small set of initial building blocks is provided, while the majority of structural refinement and prompt evolution is handled automatically.

Specifically, existing systems such as AutoGen~\cite{wu2023autogen}, GPTSwarm~\cite{Zhuge2024LanguageAA}, and ADAS~\cite{hu2025automated} also rely on manually specified templates or primitives as their starting point.
AutoGen requires developers to instantiate agents with predefined system prompts and fixed conversational patterns.
GPTSwarm initializes a hand-designed operation graph whose nodes and edges are later optimized.
ADAS searches over a code space composed of author-defined agent templates and controller patterns.
In comparison, $\mathcal{ANN}$ requires only a lightweight initialization that is comparable to, or smaller than, these prior approaches.

In practice, deploying $\mathcal{ANN}$ for a new task family only requires providing a compact JSON configuration file that defines a small number of initial candidate agent teams (typically 1--3 blocks per layer) and a few reusable agent candidates.
After this initialization step, all subsequent structural evolution, including adding or removing nodes, modifying edges, and refining prompts, is performed automatically by the LLM during the backward optimization phase.
Users are not required to manually design or maintain prompts for all future agents, which keeps the deployment cost low.

Config~\ref{config:1} shows a simplified initialization example used for the HumanEval code-review layer.
Despite its simplicity, the same initialization format is reused across diverse domains, including code generation (HumanEval), data analysis (DABench), mathematical reasoning (MATH), creative writing, knowledge-intensive question answering (MMLU--ML), and natural-language planning (Natural Plan), without redesigning the workflow from scratch.
In our experiments, the full initialization file for each dataset consists of only a few hundred lines of JSON and requires only minor domain-specific adjustments.

Table~\ref{tab:init_stats} summarizes the scale of manual initialization used across all benchmarks.
For all six datasets, the number of manually specified agent-team candidates is at most three, demonstrating that $\mathcal{ANN}$ achieves broad applicability with very limited human intervention.

\begin{table}[t]
\centering
\footnotesize
\begin{tabular}{lcc}
\hline
\textbf{Benchmark} & \textbf{\# Initial Agent Teams} & \textbf{\# Initial Agents} \\
\hline
HumanEval & 2 & 8 \\
Creative Writing & 2 & 7 \\
MATH & 3 & 7 \\
DABench & 3 & 7 \\
MMLU--ML & 2 & 8 \\
Natural Plan & 2 & 8 \\
\hline
\end{tabular}
\caption{Scale of manual initialization across benchmarks.}
\label{tab:init_stats}
\end{table}

\subsection{Prompt Repository}
\label{sec:promptexample}

To guarantee rigorous experimentation, our framework distills complex evaluation and optimisation routines into a curated suite of six reusable examples of \emph{prompts} for reference.  
Each prompt encapsulates a distinct facet of model assessment—ranging from factual exactness to strategic, multi-layer workflow repair—thereby furnishing a unified interface for loss-function design and optimiser selection.  
Collectively, these templates enable (i) \textbf{fine-grained answer verification}, (ii) \textbf{holistic workflow diagnosis}, and (iii) \textbf{progressive, momentum-aware refinement}, furnishing the gradient signals that steer the training loop towards globally coherent behaviour.
\paragraph{Answer Verification.} Prompt~\ref{prompt:1} formalises a strict comparison between a model’s predicted answer and an externally supplied ground truth, while Prompt~\ref{prompt:2} generalises the rubric to creative-writing settings where no canonical answer exists.
  \paragraph{Global Optimisation.} Prompt~\ref{prompt:3} performs gradient-based analysis over an entire workflow trajectory, isolating error-prone sub-tasks and prescribing block-level remedies.
  \paragraph{Layer-wise Repair.} Prompt~\ref{prompt:4} zooms in on a single block, recommending structural or prompt-template adjustments that preserve inter-block consistency.
  \paragraph{Momentum-based Adjustment.} Prompt~\ref{prompt:5} fuses historical “velocity” information with fresh gradient signals to resolve recurrent faults while safeguarding previously effective changes.
  \paragraph{Block Selection.} Prompt~\ref{prompt:6} scores competing blocks against task complexity, ensuring that the most capable module is invoked for code-finalisation tasks and analogous challenges.

By systematically orchestrating these prompts, we induce \emph{task-aligned} gradients that couple local correctness with global workflow efficiency, thereby enhancing both convergence speed and final performance.




\begin{tcolorbox}[colback=white, colframe=gray, title={
Prompt 1: Prompt for Answer Verification with Ground Truth }, width=\linewidth, halign=left, enhanced jigsaw, breakable]
\refstepcounter{prompt}
\label{prompt:1}
\begin{lstlisting}[language=TeX, basicstyle=\ttfamily\footnotesize, numbers=none, xleftmargin=0pt, frame=none]
You are a helpful AI assistant. 
You will use your math skills to verify the answer.

You are given:
- A problem: {problem}
- A reply from a model: {final_answer}
- A ground truth answer: {solution}

Please do the following::
1. Extract the answer from the reply in the format:
   "The answer is <answer extracted>"

2. Compare the extracted answer with the ground truth.

3. Based on your analysis, choose only one of the following outputs:
   (a) "The answer is correct."
   (b) "The answer is approximated but should be correct."
   (c) "The answer is incorrect. 
       Correct Answer: <ground truth answer> </ground truth answer> |
       Answer extracted: <answer extracted> </answer extracted>."
   (d) "The reply doesn't contain an answer."
\end{lstlisting}
\end{tcolorbox}





\begin{tcolorbox}[colback=white, colframe=gray, title={
Prompt 2: Prompt for Creative Writing Evaluation }, width=\linewidth, halign=left, enhanced jigsaw, breakable]
\refstepcounter{prompt}
\label{prompt:2}
\begin{lstlisting}[language=TeX, basicstyle=\ttfamily\footnotesize, numbers=none, xleftmargin=0pt, frame=none]
Evaluate the following creative writing piece based on the provided task.

Inputs:
- Task Description: {task_prompt}
- Creative Writing Output: {output_from_last_layer}

Evaluation Criteria:
- Logical coherence: Is the text logically organized?
- Emotional engagement: Does the text evoke the desired emotions?
- Adherence to task requirements: Does it match the original prompt?
- Creativity: Is the text original and imaginative?

Output Format:
- Coherence: [Score out of 10, with a brief explanation]
- Engagement: [Score out of 10, with a brief explanation]
- Adherence: [Score out of 10, with a brief explanation]
- Creativity: [Score out of 10, with a brief explanation]
- Suggestions for Improvement: [Text]
- Overall Score: [Score out of 10]
\end{lstlisting}
\end{tcolorbox}

\begin{tcolorbox}[colback=white, colframe=gray, title={
Config 1: Example JSON Initialization for the HumanEval Code-Review Layer}, width=\linewidth, halign=left, enhanced jigsaw, breakable]
\refstepcounter{prompt}
\label{config:1}
\begin{lstlisting}[language=json, basicstyle=\ttfamily\footnotesize, numbers=none, xleftmargin=0pt, frame=none]
{
  "blocks": {
    "CodeReviewBlock1": {
      "block_name": "CodeReviewBlock1",
      "block_id": "1",
      "block_description": "Please carefully review the following incomplete
           Python code snippet, understand its structure, logic, and existing
           functionality....",
      "block_structure_description": "Strategy selecting better solution 
           between two generated codes.",
      "block_structure_description_details": "...",
      "entry_point": "agent_pseudo_code",
      "nodes": {
        "agent_pseudo_code": {
          "agent": "agent_pseudo_code",
          "llm_model": "default_llm",
          "prompt": "..."
        },
        "agent1_review_code": {
          "agent": "agent1_review_code",
          "llm_model": "default_llm",
          "prompt": "..."
        },
        "agent2_review_code": {
          "agent": "agent2_review_code",
          "llm_model": "default_llm",
          "prompt": "..."
        },
        "agent3_decision_maker": {
          "agent": "agent3_decision_maker",
          "llm_model": "default_llm",
          "prompt": "..."
        }
      },
      "edges": [
        ["agent_pseudo_code", "agent1_review_code"],
        ["agent_pseudo_code", "agent2_review_code"],
        ["agent1_review_code", "agent3_decision_maker"],
        ["agent2_review_code", "agent3_decision_maker"]
      ],
      "end_node": "agent3_decision_maker"
    },
    "CodeReviewBlock2": {
      "block_name": "CodeReviewBlock2",
      "block_id": "2",
      "block_description": "Please carefully review the following incomplete
                  Python code snippet, understand its structure, logic, and 
                  existing functionality....",
      "block_structure_description": "Strategy generating one code as solution.",
      "block_structure_description_details": "...",
      "entry_point": "agent_review_code",
      "nodes": {
        "agent_review_code": {
          "agent": "agent_review_code",
          "prompt": "..."
        }
      },
      "edges": [],
      "end_node": "agent_review_code"
    }
  },
  "nodes": {
    "agent_review_code_after_pseudo_code": {
      "agent": "agent_review_code",
      "prompt": "..."
    },
    "agent_decision_maker_with_2_options": {
      "agent": "agent_decision_maker_with_2_options",
      "prompt": "..."
    },
    "agent_static_analysis": {
      "agent": "agent_static_analysis",
      "prompt": "..."
    }
  }
}
\end{lstlisting}
\end{tcolorbox}

\begin{tcolorbox}[colback=white, colframe=gray, title={
Prompt 3: Prompt for Gradient-Based Global Optimization  }, width=\linewidth, halign=left, enhanced jigsaw, breakable]
\refstepcounter{prompt}
\label{prompt:3}
\begin{lstlisting}[language=TeX, basicstyle=\ttfamily\footnotesize, numbers=none, xleftmargin=0pt, frame=none]
Task:
You are an advanced global workflow analysis assistant tasked with diagnosing 
inefficiencies and proposing optimizations for a multi-step process. 
Your goal is to analyze the workflow trajectory and determine which aspects
need improvement to address task failures and enhance overall performance.

You will evaluate the provided consolidated information from a workflow task. 
Identify which sub-task outputs or prompts likely caused the failure and
provide specific suggestions for each sub-task. Your output must strictly
follow this format: 
<output\_format>\{example\_global\_loss\_format\}</output\_format>

Important Notice:
- All analyses and suggestions should be based on a general level.
- Avoid overly targeted feedback for this specific task instance.
- All required information is provided via: {initial_solution}

Global Optimization Steps:

1. Final Result Evaluation:
   Analyze the final result <final result> to determine if the task failed.

2. Solution Comparison:
   Compare <canonical solution> and <generated solution>:
   - Is the logic in <generated solution> aligned with <canonical solution>?
   - Where is the gap between the analysis and the standard answer?
   - Identify specific issues in <generated solution> that contributed to the
     failure.
   - Document these findings in the 'global_analysis' section of the  
     <output_format>.

3. Block Input and Output Analysis:
   Based on the <task description> and <workflow trajectory>:
   - Do not compare the block outputs with the <canonical solution>.
   - Examine each block_input and block_output.
   - Identify which block(s) caused the task to fail.
   - Highlight any inefficiencies or redundancies.
   - Write optimization suggestions into the 'structure_suggestion' section
     of each relevant block.
   - Review each block's block_description and provide edits if necessary,
     recorded in the 'prompt_suggestions' section.
   - If no edits are needed, do not add any suggestions.

4. Node-Level Analysis Within Blocks:
   For each problematic block:
   - Analyze the internal node_input and node_output.
   - Evaluate the team collaboration structure.
   - Propose improvements to intra-block agent collaboration, if necessary.
   - Document your suggestions in the 'structure_suggestion' section of the
     corresponding block.

\end{lstlisting}
\end{tcolorbox}

\begin{tcolorbox}[colback=white, colframe=gray, title={
Prompt 4: Prompt for Layer-Wise Block Optimization}, width=\linewidth, halign=left, enhanced jigsaw, breakable]
\refstepcounter{prompt}
\label{prompt:4}
\begin{lstlisting}[language=TeX, basicstyle=\ttfamily\footnotesize, numbers=none, xleftmargin=0pt, frame=none]
You are given a block within a workflow. Your task is to suggest 
optimizations for this block, focusing on both prompt improvements and 
structural changes, while ensuring consistency and efficiency.

Block Information:
- Block Name: {block_name}
- Global Loss Feedback: {global_loss_feedback}
  (This is global feedback for the entire workflow. Use as reference, but 
  base suggestions on block-level reasoning.)
- Blocks Log: {blocks_log}
  (Includes architecture, node inputs/outputs, block/node descriptions.)
- Canonical Solution: {canonical_solution}
- Task Description: {task_prompt}

Evaluation Criteria:
1. Evaluate Each Node
- Check input_variables for validity and consistency.
- Valid sources include:
  * State variables: "task_data", "task_prompt", "task_id"
  * Prior node outputs: e.g., calculation_expert1_output
- For prompt modifications:
  * Include an updated prompt_template with clear instructions
  * Explicitly list all input_variables and their sources

2. Propose Structural Changes
- Add/remove nodes (max 3 additions)
- For added nodes, specify:
  * node_name, agent, output format, prompt_template
  * variable_sources, constraints
- Define from/to edges for new nodes
- Update connected nodes' input_variables if needed
- Set the new entry_node and end_node
- Ensure all nodes (except end_node) have valid outgoing edges
- Include all_edges_now and all_nodes_now

3. Impact on Other Nodes
- Maintain logical consistency with the entire workflow

4. Use Available Agents
- Refer to {available_agents} for potential agents
- Check each agent's constraints for fit
- Modify agents as needed (update prompt_template, input_variables, or define
new agents)

5. Dynamic Block ID and Naming
- Use {new_block_id} to assign a unique block_id
- Format name as {block_name}X, where X = new_block_id

6. Block Structure Description
- Include:
  * block_structure_description: high-level purpose
  * block_structure_description_details: including:
    1. Nodes and connections
    2. Node roles and logic
    3. Input/output flow
- Ensure clarity, accuracy, and alignment with structure

7. Provided Canonical Solution and Test Cases
- Don't over-optimize: block may not be the cause of failure
- Avoid overfitting: feedback should remain generalized
- Use <canonical solution> and <test cases> as reference only

8. Output Format
- All feedback must be returned in this JSON format: {layerwise_loss_format}
- Do not use arrows to represent edges!
\end{lstlisting}
\end{tcolorbox}

\begin{tcolorbox}[colback=white, colframe=gray, title={
Prompt 5: Prompt for Momentum-Based Adjustment}, width=\linewidth, halign=left, enhanced jigsaw, breakable]
\refstepcounter{prompt}
\label{prompt:5}
\begin{lstlisting}[language=TeX, basicstyle=\ttfamily\footnotesize, numbers=none, xleftmargin=0pt, frame=none]
Task Description:
You are an advanced strategic advisor focused on enhancing team performance.
Your role is to analyze recent feedback in combination with historical 
adjustments to guide team improvement for a specific workflow block.

Provided Information:
- Team Name: <team name> {block_name} </team name>
- Current Team Structure: <current team> {current_block} </current team>
- Final Result of Task Execution:
  <final result> {current_task_results} </final result>
- Current Gradient Feedback: 
  <current feedback> {current_gradient} </current feedback>
- Previous Adjustment Direction: 
  <previous adjustment direction> {velocity}</previous adjustment direction>
- Input and Output for Block and Nodes:
  * <team input> {block_input} </team input>
  * <team output> {block_output} </team output>
  * <input and output of all nodes> {nodes_info} 
    </input and output of all nodes>

Instructions - Two-Step Strategy:

1. Overlap Handling:
- If <current feedback> overlaps with </previous adjustment direction>, focus
on these overlapping issues.
- Since the current version <current team> was formed via previous 
adjustments, but <final result> still failed, analyze why earlier suggestions
did not work.
- Carefully review block_input, block_output, and nodes_info to pinpoint 
reasons for failure.
- Revise the <current feedback> so it addresses overlapping issues in a more 
effective way.

2. New Issues Maintenance:
- If <current feedback> introduces new problems not found in 
  <previous adjustment direction>, retain those.
- Slightly refine and consolidate all suggestions to form an updated version
  of feedback.

Important Notes:
- This block may not be the root cause of task failure. 
  Avoid over-optimization.
- Our optimization is dataset-level, not task-specific. Do not overfit 
  feedback to this task instance.

Output Format:
Return your suggestions using the same structure as <current feedback>, 
wrapped as:
<adjusted feedback> [Your updated suggestions here] </adjusted feedback>
\end{lstlisting}
\end{tcolorbox}

\begin{tcolorbox}[colback=white, colframe=gray, title={
Prompt 6: Prompt Example for Layer Selection Based on Task Difficulty}, width=\linewidth, halign=left, enhanced jigsaw, breakable]
\refstepcounter{prompt}
\label{prompt:6}
\begin{lstlisting}[language=TeX, basicstyle=\ttfamily\footnotesize, numbers=none, xleftmargin=0pt, frame=none]
Task Description:
You are a performance evaluator tasked with selecting the most suitable block
for solving a Python code finalization task.  
The complete workflow consists of three blocks: code_review_block, 
code_finalize_block, and code_execute_block.

Current Block:
The block under evaluation is code_finalize_block, which represents the second
layer in the workflow.  
It's purpose is to refine another agent's code output based on prior messages, 
considering:
- Syntax accuracy
- Logical completeness
- Adherence to the initial coding intent

If the code meets the above standards, keep it unchanged. Otherwise, provide a 
corrected version.

Task Details:
- Task Objective: Improve the agent's output code using the contextual 
messages.
- Task Description: <task description> {task_prompt} </task description>

Available Blocks:
Below is a list of available blocks, including their structural roles and 
descriptions:
<list of all block's structure description> {blocks_structure_descriptions} 
</list of all block's structure description>

Instructions:
1. Evaluate the <task description> carefully, identifying key difficulty 
points and requirements.
2. Compare block roles and structures from <list of all block's structure 
description> to determine which best fits the task.
3. Select the most appropriate block based on the task complexity.

Output Format:
- Output your selection using the exact format below:
  <selected_agg_func> X </selected_agg_func>
- For example, selecting CodeFinalizeBlock3 should result in:
  <selected_agg_func> 3 </selected_agg_func>
\end{lstlisting}
\end{tcolorbox}





\section{Experiment Details}
\label{sec:details}

\subsection{Training and Validation Split}

We adopt an 80/20 protocol per dataset when comparable to prior work, and otherwise follow the dataset-specific setups cited below. This keeps splits aligned with baselines while ensuring evaluation on unseen tasks. The table fuses all details (counts, protocol, and per-task runtime).

\begin{table}[h]
\centering
\small
\setlength{\tabcolsep}{6pt}
\renewcommand{\arraystretch}{1.15}
\begin{tabular}{lrrrrcc}
\hline
\textbf{Benchmark} & \textbf{Total} & \textbf{Train} & \textbf{Val} & \textbf{Split} & \textbf{Train / task} & \textbf{Test / task} \\
\hline
HumanEval$^{\mathrm{a}}$           & 164 & 131 & 33  & 80/20  & 7 s $\uparrow$ 30 s  & 7 s  \\
Creative Writing$^{\mathrm{a}}$    & 100 &  80 & 20  & 80/20  & 17 s $\uparrow$ 30 s & 17 s \\
MATH (subset)$^{\mathrm{b}}$       & 196 & 157 & 39  & 80/20  & 13 s $\uparrow$ 33 s & 13 s \\
DAbench$^{\mathrm{c}}$             & 257 & 206 & 51  & 80/20  & 15 s $\uparrow$ 34 s & 15 s \\
MMLU--ML$^{\mathrm{d}}$ & 128 &  16 & 112 & custom & 8 s $\uparrow$ 28 s  & 8 s  \\
\hline
\end{tabular}
\caption{Dataset splits and approximate per-task runtimes. $\uparrow$ indicates additional time from backward optimization during training; inference is forward-only. 
Notes: (a) 8:2 as in \citet{zhou2024symbolic}; (b) 196-problem subset and sampling as in \citet{song2024adaptiveinconversationteambuilding}; (c) full dataset with 8:2; (d) ML subset as in \textsc{TextGrad}~\cite{yuksekgonul2024textgrad}, base dataset \citet{hendrycks2021measuringmassivemultitasklanguage}.}
\end{table}

\subsection{Effect of Training Data Size}
\label{sec:exp_training_fraction}

To further quantify the generalization behavior of $\mathcal{ANN}$ and evaluate its sensitivity to the amount of available training data, we conduct an additional analysis under different training regimes.
All experiments in this subsection follow the dataset-level optimization protocol described in Appendix~\ref{sec:case_generalization} and are evaluated exclusively on held-out test tasks that are never used during textual backpropagation.

Specifically, we consider three training settings for each benchmark:
\textbf{(i) No Train}, where $\mathcal{ANN}$ executes the initial forward-only workflow without any backward optimization;
\textbf{(ii) 50\% Train}, where textual backpropagation is performed using only half of the available training tasks; and
\textbf{(iii) Full Train}, where the framework is optimized using the complete training split.
In all cases, evaluation is conducted on the same unseen test set.
All results reported in this subsection are obtained using GPT-4o-mini to ensure consistency.

Table~\ref{tab:training_fraction} summarizes the results across HumanEval, Creative Writing, MATH, and DABench.
We report the total number of tasks, the sizes of the training and test splits, and the corresponding test performance under each training regime.
In addition, the table reports the growth of candidate agent team pools across layers, illustrating how textual backpropagation enriches the set of available agent teams during training.

Overall, the results show that textual backpropagation consistently improves test-time performance over the forward-only baseline, even when applied to a reduced training set.
Training with 50\% of the available tasks already yields noticeable gains over the untrained configuration, while full training further enhances performance.
These trends indicate that $\mathcal{ANN}$ learns reusable agentic structures that generalize beyond the specific training instances, rather than relying on per-task re-optimization.

\begin{table}[t]
\centering
\footnotesize
\setlength{\tabcolsep}{4pt}
\renewcommand{\arraystretch}{1.3}
\begin{tabular}{l c c c c c c}
\hline
\textbf{Benchmark} &
\textbf{Total} &
\textbf{Train} &
\textbf{Test} &
\textbf{Results (No Train)} &
\textbf{Results (50\% Train)} &
\textbf{Results (Full Train)} \\
\hline
HumanEval        & 164 & 131 & 33 & 86.8 & 87.9 & 90.9 \\
Creative Writing & 100 &  80 & 20 &  8.3 &  8.5 &  9.0 \\
MATH             & 196 & 157 & 39 & 65.0 & 70.0 & 82.5 \\
DABench          & 257 & 206 & 51 & 86.3 & 86.3 & 90.2 \\
\hline
\end{tabular}
\caption{
Test performance under different training regimes.
``No Train'' denotes forward-only execution without textual backpropagation;
``50\% Train'' denotes optimization using half of the available training tasks;
``Full Train'' denotes optimization using the complete training split.
All results are evaluated on unseen test tasks and use GPT-4o-mini.
}
\label{tab:training_fraction}
\end{table}

\section{Case Study}
\subsection{Prompt Evolutions Examples}
\label{sec:promptchange}

\begin{figure*}[t]
\centering
  \includegraphics[width=0.9\linewidth]{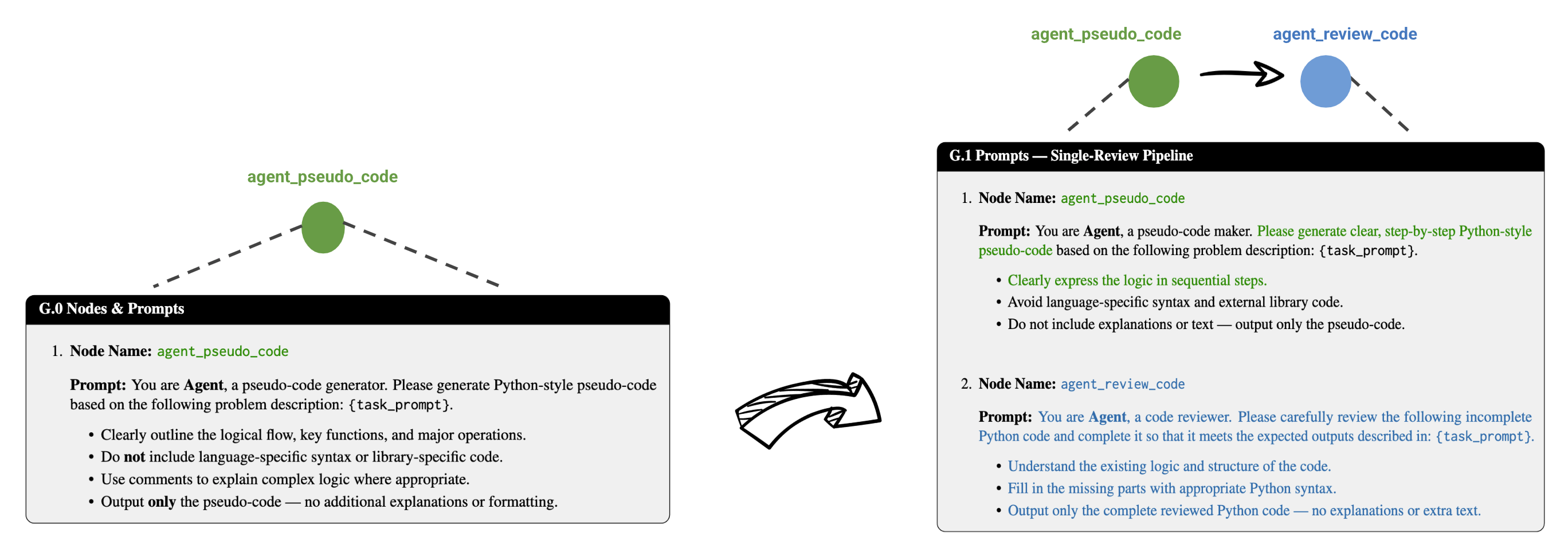}\hfill
  \includegraphics[width=0.9\linewidth]{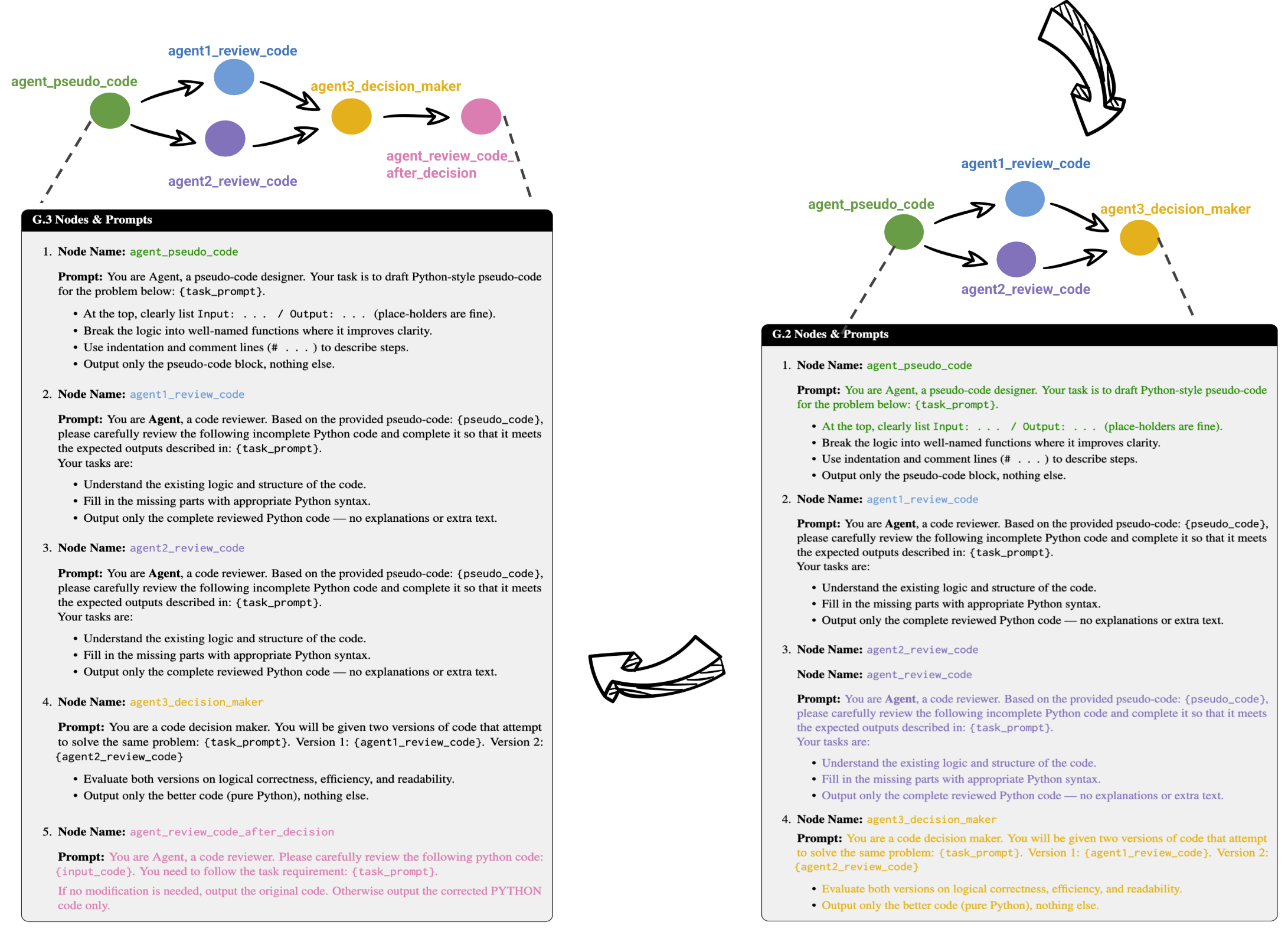}\hfill
  \includegraphics[width=0.5\linewidth]{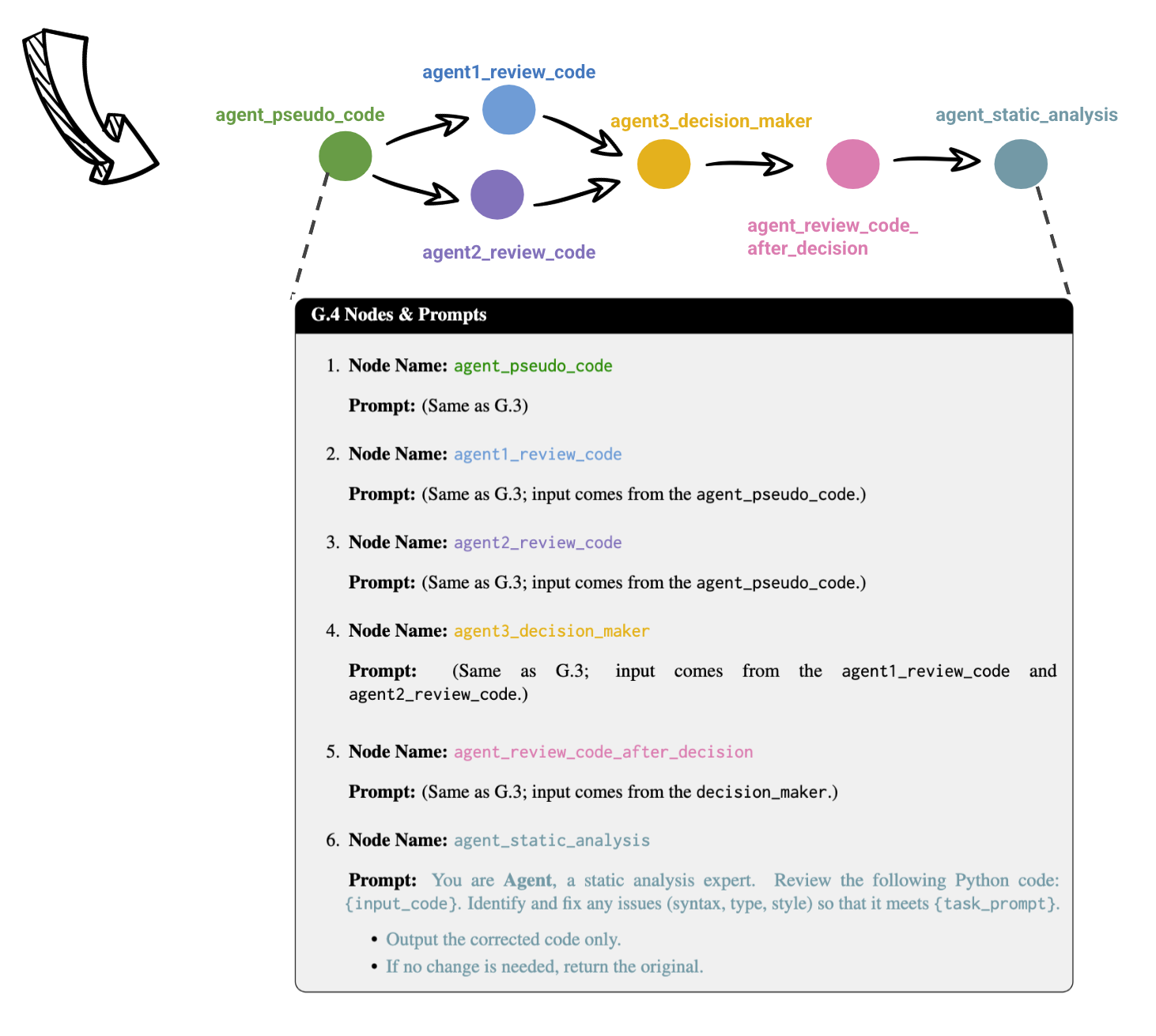}\hfill
\caption{%
    Prompt-evolution trajectory for the \texttt{HumanEval}\cite{chen2021evaluatinglargelanguagemodels} \textit{review\_code} subtask. Boxes denote agent nodes, arrows indicate information flow, and shaded regions highlight components newly introduced at each iteration.%
  }
\label{fig:humaneval_prompt_change_plot}
\end{figure*}

\begin{figure*}[t]

  \includegraphics[width=0.6\linewidth]{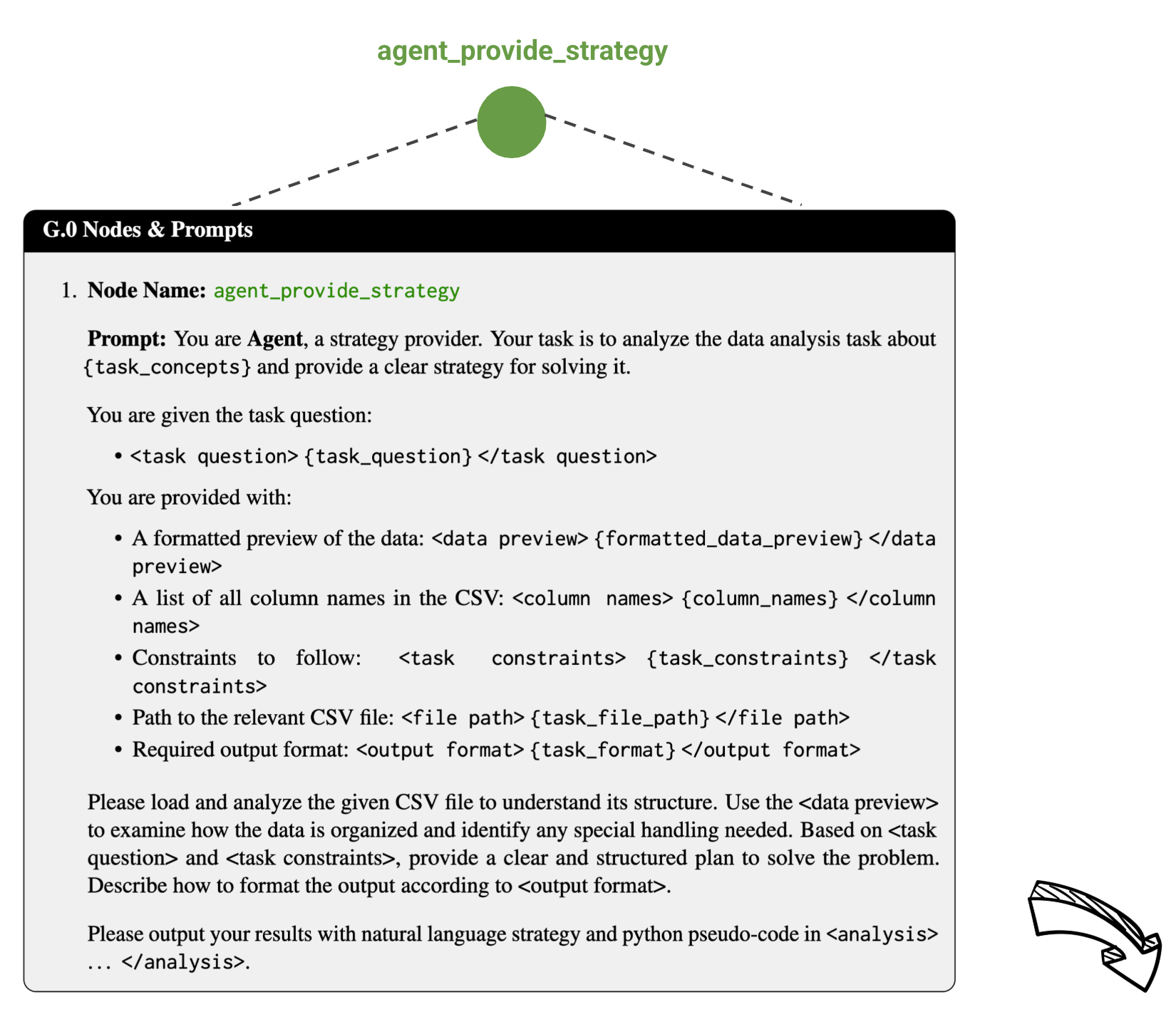}\hfill\\
  \includegraphics[width=1\linewidth]{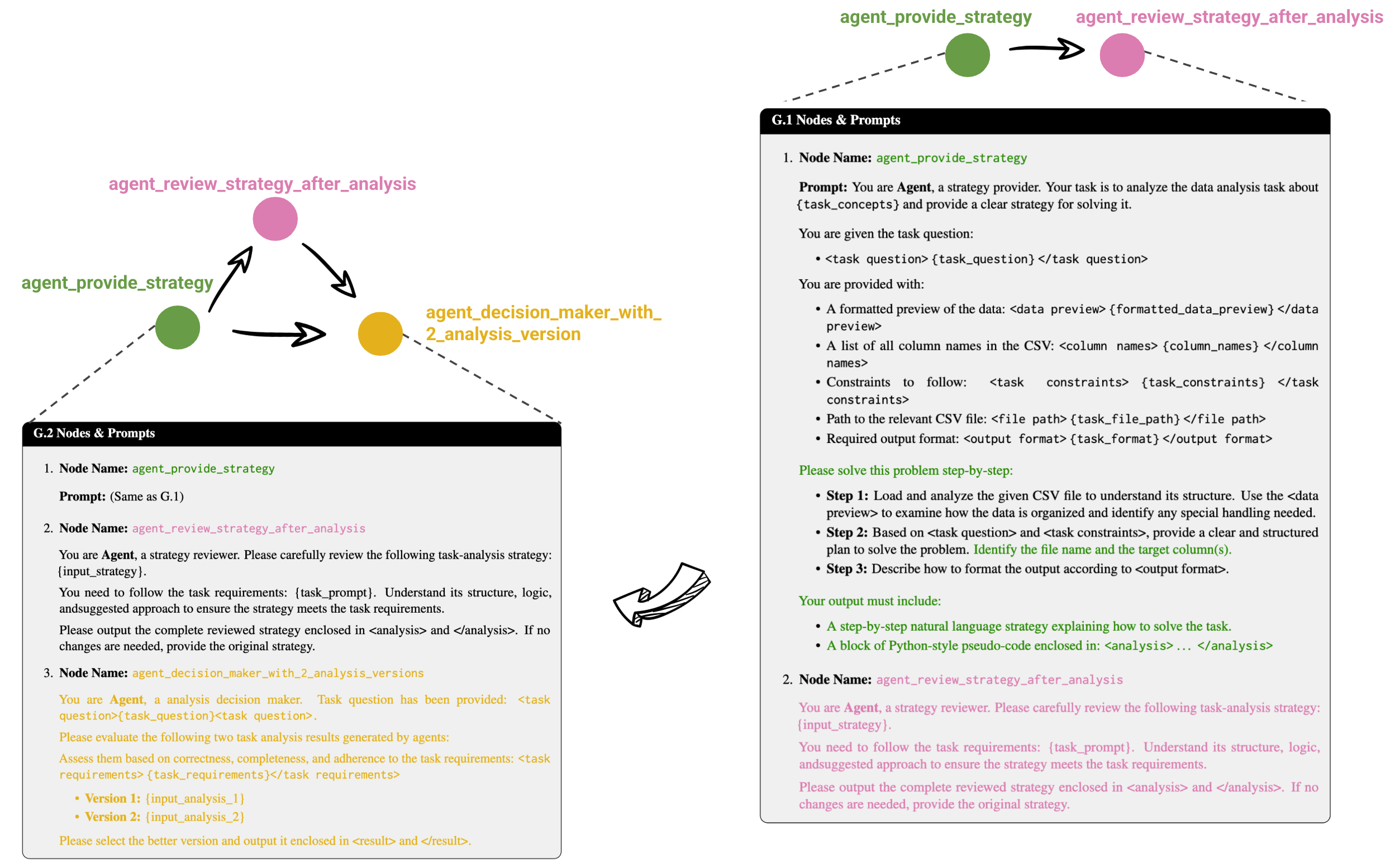}\hfill

\caption{
Prompt-evolution trajectory for the \texttt{DABench}\cite{Hu2024InfiAgentDABenchEA} task-analysis benchmark. Boxes denote agent nodes, arrows indicate information flow, and shaded regions highlight components newly introduced at each iteration.
}
\label{fig:dabench_prompt_change_plot}
\end{figure*}

Figure~\ref{fig:humaneval_prompt_change_plot} and Figure~\ref{fig:dabench_prompt_change_plot} illustrate representative trajectories of prompt evolution across two benchmark tasks: subtask about code review in the HumanEval dataset and subtask about task analysis in the DABench suite, respectively. These diagrams reflect both the structural transformations of block-level workflows and the fine-grained progression of node-level prompt design. Together, these visualizations exemplify how the prompt design co-evolved with structural modularity. 

\paragraph{HumanEval: Code Review Prompt Evolution.}
Figure~\ref{fig:humaneval_prompt_change_plot} demonstrates how the system's prompt architecture evolved in the context of solving the \texttt{review\_code} subtask on the HumanEval dataset. Initially, the workflow consisted of a single-agent node responsible for completing partially written code. As the system matured, this simplistic design was incrementally augmented with a multi-agent framework involving two parallel reviewers and a subsequent decision node. Each reviewer agent received increasingly structured prompts, incorporating pseudo-code context, explicit reasoning criteria (e.g., correctness, efficiency, readability), and modular output constraints.

In subsequent iterations, the system integrated static analysis agents, forming a pluggable review-correction pipeline. The final prompt configuration emphasized modular roles, strict output formatting, and conditional rewriting policies, resulting in a robust, interpretable code review pipeline.

\paragraph{DABench: Task Analysis Prompt Evolution.}
Figure~\ref{fig:dabench_prompt_change_plot} illustrates the evolution of task analysis prompts when solving data-centric reasoning problems in the DABench benchmark. The initial system was anchored around a single agent generating a natural-language strategy and accompanying pseudo-code. Prompt instructions were general-purpose, with minimal context sensitivity or structural annotation.

With successive iterations, the system adopted a multi-agent architecture, introducing review, feedback, and revision loops. Each agent’s prompt was incrementally specialized: reviewers were instructed to analyze structural logic, adherence to constraints, and planning completeness. Prompts began incorporating input-specific metadata, including task constraints, file paths, and structured output tags (e.g., \texttt{<analysis>}, \texttt{<feedback>}, \texttt{<result>}).

\subsection{Team Structure Examples with Optimization}
\label{sec:teamexample}

To better understand how agent team structures evolve throughout the optimization process, we present visualizations of team configurations across multiple datasets. These examples demonstrate how architectures transition from simple, linear pipelines to more dynamic, graph-based systems as the model learns to coordinate more effectively.

\begin{figure*}[t]
    \centering
    \input{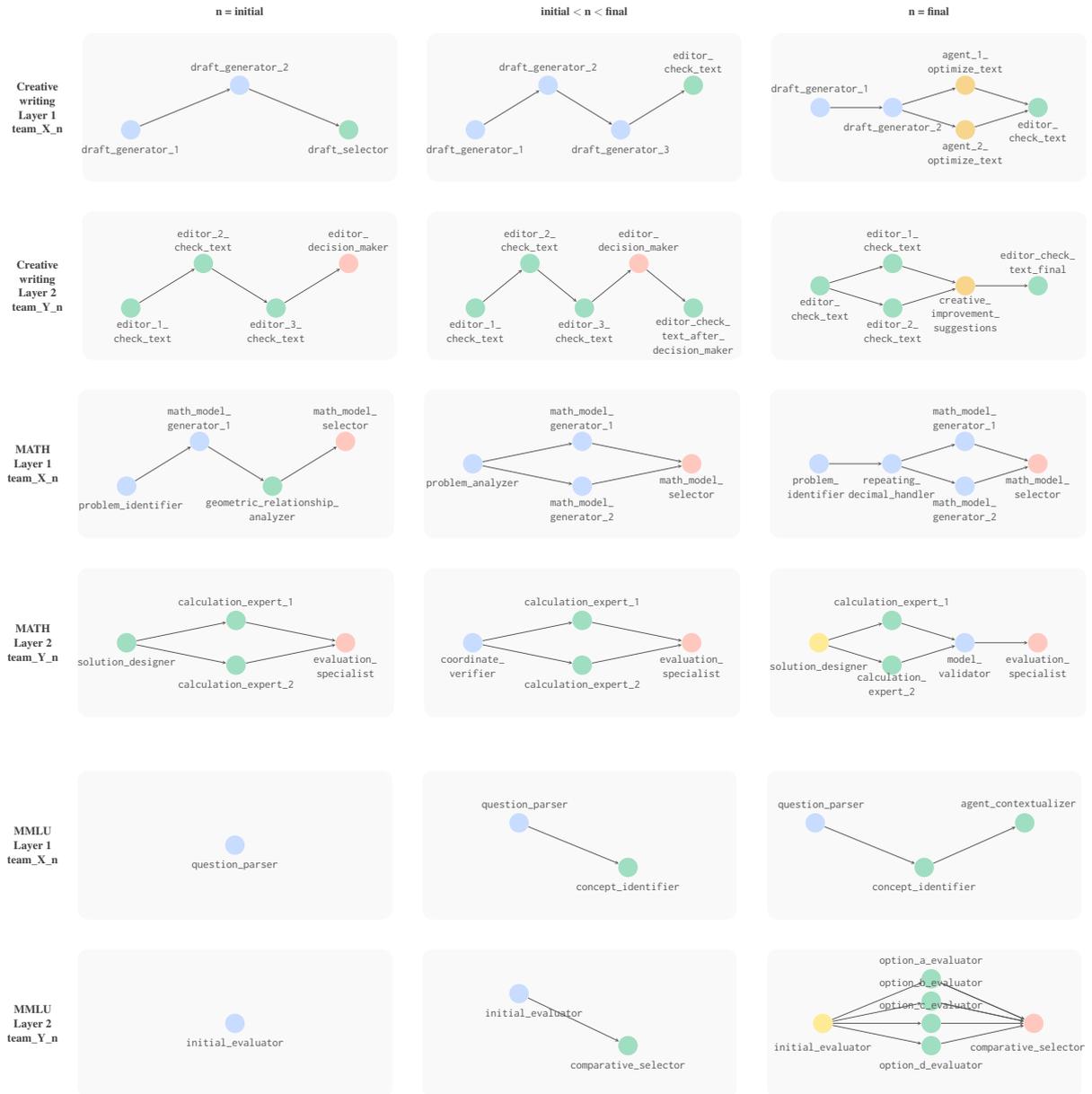} 
    \caption{Evolution of agent team structures on the \textbf{Creative Writing}~\cite{zhou2024symbolic}, \textbf{Math}~\cite{hendrycks2021measuringmathematicalproblemsolving}, and \textbf{MMLU–Machine Learning}~\cite{hendrycks2021measuringmassivemultitasklanguage} datasets. For each dataset, we visualize a representative example from one layer, showing how team configurations become progressively more structured and cooperative through optimization.}
    \label{fig:structure_cw_math_mmlu}
\end{figure*}

Figure~\ref{fig:structure_cw_math_mmlu} illustrates selected examples from three representative datasets: Creative Writing~\cite{zhou2024symbolic}, Math~\cite{hendrycks2021measuringmathematicalproblemsolving}, and MMLU–Machine Learning~\cite{hendrycks2021measuringmassivemultitasklanguage}. For each dataset, we choose a single layer and show how the team structure at that layer evolves over time. As optimization progresses, the agent configurations become increasingly complex and tailored to the demands of each dataset, reflecting greater specialization and improved collaboration.

\begin{figure*}[t]
    \centering
    \input{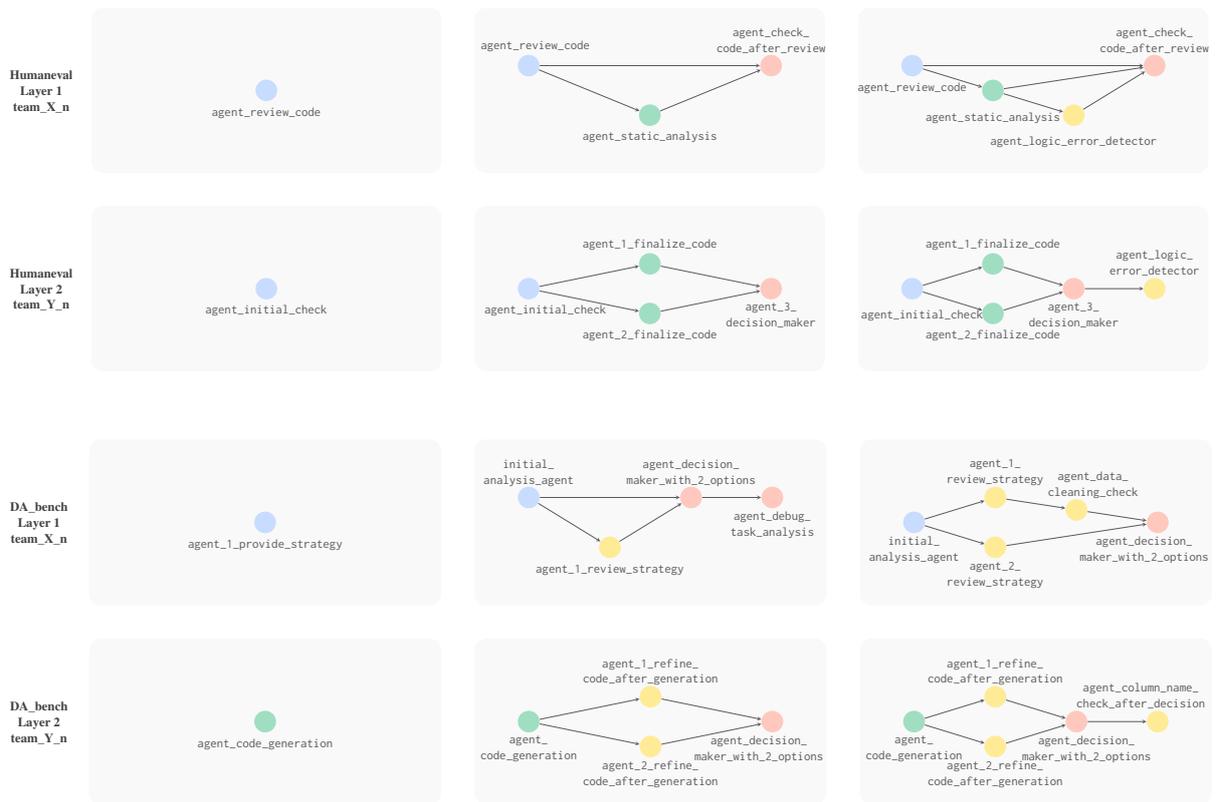} 
    \caption{Team structure visualizations for the \textbf{HumanEval}~\cite{chen2021evaluatinglargelanguagemodels} and \textbf{DABench}~\cite{Hu2024InfiAgentDABenchEA} datasets. Each node's color reflects its functional role within the system. The diagrams highlight how different types of agents coordinate and how task-specific configurations emerge over time.}
    \label{fig:structure_humaneval_dabench}
\end{figure*}

Figure~\ref{fig:structure_humaneval_dabench} focuses on two additional datasets: HumanEval~\cite{chen2021evaluatinglargelanguagemodels} and DABench~\cite{Hu2024InfiAgentDABenchEA}. In the case of DABench, we adopt the random train/validation split from~\cite{song2024adaptiveinconversationteambuilding}. Here, we emphasize the functional diversity among agents by using different node colors to indicate distinct roles (e.g., generation, evaluation, decision-making). These visualizations highlight how functional heterogeneity and task-specific routing emerge through optimization.

Together, these figures demonstrate how adaptive reconfiguration of agent teams enables more effective problem solving and reflects the system’s ability to internalize dataset-specific strategies.

\subsection{Stability and Failure Modes}

Text-based optimization with large language models can introduce noise, hallucination, or drift, especially when structural updates are generated in free-form natural language.
While the main body of this paper focuses on successful optimization trajectories, it is equally important to understand when and how textual backpropagation fails, and how such failures are prevented from propagating across layers.
In $\mathcal{ANN}$, stability is explicitly enforced through a combination of momentum-based updates and a multi-stage validation mechanism that filters out unstable, ill-formed, or non-improving updates before they are integrated into the candidate pool.

\paragraph{Momentum-based stabilization.}
Beyond validation, $\mathcal{ANN}$ employs momentum to stabilize structural optimization across iterations.
Rather than treating each textual gradient independently, momentum maintains a directional memory of previous update steps, encoding how and why an agent team was modified in earlier iterations.
This historical trajectory is integrated with newly computed textual gradients, which helps suppress oscillatory updates and enables error-aware correction when previous optimizations fail.
As a result, structural changes evolve smoothly over time instead of reacting abruptly to noisy feedback.

\paragraph{Multi-stage validation and error interception.}
As summarized in Algorithm~2, every locally updated agentic workflow must pass a sequence of validation checks before it is accepted.
These include \emph{node validation} (e.g., \textsc{VariableSourcesValid}, \textsc{FormatValid}), \emph{edge validation} (e.g., \textsc{AllNodesHaveEdges}), \emph{structure validation} (e.g., \textsc{StructureNotUnique}), and \emph{performance validation}.
Only candidate blocks that satisfy all validation criteria are allowed to enter the candidate pool and influence subsequent layers.

In practice, a non-trivial fraction of LLM-generated updates are rejected by these checks.
This rejection mechanism plays a crucial role in preventing hallucinated structures, invalid variable dependencies, or redundant architectures from propagating errors through the layered workflow.
Table~\ref{tab:failure_cases} presents representative failure cases observed during training, illustrating how different validation stages intercept problematic updates.
For example, node validation can reject updates that introduce invalid variable sources, structure validation can prevent duplicate block configurations, and performance validation can filter out changes that do not yield measurable improvements.

Taken together, these mechanisms ensure that textual backpropagation in $\mathcal{ANN}$ is not a blind acceptance of LLM suggestions.
Instead, optimization is tightly constrained by explicit validation rules, performance-based filtering, and momentum-based smoothing, ensuring that errors remain localized and do not cascade across layers.
Consistent with this design, our ablation studies show that removing momentum or validation components leads to noticeably less stable optimization behavior.

\begin{table}[t]
\centering
\footnotesize
\setlength{\tabcolsep}{4pt}
\renewcommand{\arraystretch}{1.3}
\begin{tabular}{l l l p{5cm} p{3.0cm}}
\hline
\textbf{Validation type} & \textbf{Task ID} & \textbf{Block (candidate)} & \textbf{Failure reason} & \textbf{Log summary} \\
\hline

\textbf{\makecell[l]{Node\\Validation}} &
\makecell[l]{HumanEval\\/67} &
CodeReviewBlock3 &
Invalid variable source assignment: node \texttt{agent\_input\_validation} sets \texttt{variable\_sources["input\_string"]} to an unregistered source (\texttt{node\_input}). &
Invalid variable source detected; update rejected. \\

\textbf{\makecell[l]{Structure\\Validation}} &
\makecell[l]{HumanEval\\/127} &
CodeFinalizeBlock6 &
Duplicate structure detected: edge set \texttt{all\_edges\_now} matches an existing block (\texttt{CodeFinalizeBlock4}). &
Duplicate block structure; update rejected. \\

\textbf{\makecell[l]{Performance\\Validation}} &
\makecell[l]{HumanEval\\/145 }&
CodeFinalizeBlock8 &
No performance gain: \texttt{validate\_performance} returns \texttt{False} despite passing structural checks. &
Predicted performance not improved; update rejected. \\

\hline
\end{tabular}
\caption{Representative failure cases intercepted by different validation stages during textual backpropagation.}
\label{tab:failure_cases}
\end{table}

\subsection{Dataset-Level Optimization and Generalization}
\label{sec:case_generalization}

A key design principle of $\mathcal{ANN}$ is the explicit separation between \emph{offline dataset-level optimization} and \emph{online test-time execution}.
Rather than re-optimizing the system for each individual task instance, our goal is to learn reusable agent teams for a \emph{task family} (e.g., HumanEval coding problems, MATH competition questions, or DABench data-analysis tasks) and then apply the learned architecture to unseen tasks from the same distribution.

Concretely, for each benchmark, $\mathcal{ANN}$ is trained once using a designated training split.
During this phase, textual backpropagation is applied only to training tasks: execution trajectories from multiple tasks are aggregated, and both global and local textual gradients are used to update candidate agent teams, prompts, and inter-layer connections.
No backward optimization is performed on validation or test examples.
After training, the resulting candidate team pools and workflow structure are frozen and reused for all subsequent test-time evaluations.

At inference time, $\mathcal{ANN}$ operates in a purely forward manner.
Each layer maintains a fixed candidate pool of agent teams, and a team-selector agent dynamically routes the input by selecting the most suitable team based on the current subtask and execution context.
Importantly, no further textual gradients, architectural updates, or prompt modifications are performed during evaluation.
As a result, test-time performance reflects the generalization capability of the learned agentic architecture rather than per-instance adaptation.

This protocol is consistently applied across all benchmarks.
For HumanEval, Creative Writing, MATH, and DABench, we adopt an 80\%/20\% train--test split, while for MMLU--ML we follow the dataset-specific split used in prior work.
All results reported in the main paper are obtained by optimizing $\mathcal{ANN}$ on the training split and evaluating on disjoint, unseen test tasks.
Quantitative evidence under different training regimes, including forward-only execution and partial training, is reported in Appendix~\ref{sec:exp_training_fraction}.

\end{document}